\documentclass[sigconf,table,dvipsnames,nonacm]{acmart}

\usepackage{booktabs}
\usepackage{array}
\usepackage{placeins} 

\DeclareUnicodeCharacter{2264}{\ensuremath{\leq}}

\renewcommand\footnotetextcopyrightpermission[1]{} 

\AtBeginDocument{%
  \providecommand\BibTeX{{%
    \normalfont B\kern-0.5em{\scshape i\kern-0.25em b}\kern-0.8em\TeX}}}

\begin{document}

\title{Evaluating the Robustness of Anti-UAV Detection under
Controlled Fog Degradation}
\subtitle{Fog-Aware Training and Clear-Sky Tradeoff}

\author{Gur Levy Birkental}
\affiliation{%
  \institution{University of Amsterdam}
  \city{Amsterdam}
  \country{The Netherlands}
}
\email{gur.levy@student.uva.nl}

\author{Seyed Sahand Mohammadi Ziabari}
\authornote{Corresponding author.}
\affiliation{%
  \institution{University of Amsterdam}
  \city{Amsterdam}
  \country{The Netherlands}
}
\affiliation{%
  \institution{Department of Computer Science and Technology, SUNY Empire State University}
  \city{Saratoga Springs}
  \state{NY}
  \country{USA}
}
\email{s.s.mohammadiziabari@uva.nl}

\author{Ali Mohammed Mansoor Alsahag}
\affiliation{%
  \institution{University of Amsterdam}
  \city{Amsterdam}
  \country{The Netherlands}
}
\email{a.m.m.alsahag@uva.nl}

\begin{abstract}

{
Vision-based anti-UAV systems must function in poor visibility, yet most benchmarks use only clear-sky footage, and previous robustness studies treat adverse weather as a simple present/absent condition. As a result, the impact of fog severity on ground-to-air UAV detection remains poorly understood. This work presents the first severity-controlled fog benchmark for this task: synthetic fog at ten severity levels is applied to the RGB modality of the Anti-UAV300 dataset, comparing a clear-trained YOLOv5m baseline to a fog-aware model trained on both clear and foggy images. Detection performance drops sharply and non-linearly: degradation is front-loaded across light-to-moderate fog ($\beta \approx0.05-0.10$), with a $96\%$ reduction in mAP@0.5:0.95 from clear to thickest fog, mainly due to lost recall and confidence. On the comparable metric (mAP@0.5), this collapse exceeds the most extreme rain degradation reported in the closest prior benchmark. Fog-aware training boosts detection across all severities (up to $+0.320$ mAP@0.5:0.95) with only a $9.1\%$ drop in clear-sky accuracy, raising the threshold for reliable detection while not preventing collapse under extreme fog. Since reliability is lost within a narrow visibility range, simple clear-versus-adverse tests underestimate operational risk. 
}

\end{abstract}



\fancyhead{}
\maketitle
\section{Introduction}
\label{sec:introduction}
Object tracking is one of the most important and innovative tasks in computer vision, aiming to detect and track objects across image sequences \cite{soleimanitaleb_object_2019, smeulders_visual_2014}. This task has many applications including video surveillance \cite{zhao_fine-grained_2020, jha_real_2021}, traffic monitoring \cite{ghahremannezhad_object_2023, el-alami_review_2024}, and even medical diagnosis systems \cite{albuquerque_deep_2025,ragab_comprehensive_2024}. 
The growing importance of object tracking is particularly evident in the context of Unmanned Aerial Vehicles (UAVs) \cite{wang_progress_2025, jiang_anti-uav_2023, huang_anti-uav410_2024, ding_vision-based_2025}. UAV accessibility and popularity have transformed them from recreational novelties into significant instruments for surveillance and monitoring; however, their misuse now poses escalating risks to public safety and personal privacy \cite{aldosari_drone-type-set_2024,coluccia_drone-vs-bird_2024,cubber_standardized_2025,ding_vision-based_2025}. In safety-critical surveillance contexts, unauthorized UAV incursions present severe threats: sightings at airports have repeatedly halted air traffic, leading to massive economic losses and significant operational disruptions \cite{ding_vision-based_2025,jiang_anti-uav_2023,aldosari_drone-type-set_2024,cubber_standardized_2025}. 

A prominent illustration of this vulnerability occurred during the 36-hour closure of London Gatwick Airport, where over 1000 flights were canceled following multiple drone sightings. Reports from this incident indicated that dense fog and light precipitation significantly reduced the effectiveness of visual and thermal cameras \cite{hudson_gatwick_nodate}. Because poor visibility prevented consistent tracking, law enforcement was unable to identify the culprit or confirm whether the threat was still present. 

This incident highlights a reliability gap in existing vision-based detection algorithms when deployed in high-stakes environments \cite{munir_impact_2024, mudavath_object_2025, tahir_object_2024, appiah_object_2024}. Most existing benchmarks are constructed under clear atmospheric conditions, failing to account for the uncontrollable weather artifacts that safety-critical systems must endure to provide 24/7 protection\cite{ding_vision-based_2025,feng_hazydet_2025,munir_impact_2024,xie_cst_2025}. For instance, existing large-scale anti-UAV benchmarks, despite comprising hundreds of thousands of manually annotated bounding boxes across hundreds of video sequences, contain only clear-sky conditions \cite{jiang_anti-uav_2023, huang_anti-uav410_2024, xie_cst_2025}. Recent studies confirm that even top-tier detectors suffer performance degradation, when confronted with rain streaks, fog, or motion blur\cite{munir_impact_2024, agarwal_warlearn_2024,feng_hazydet_2025}.

However, while the sensitivity of detectors to adverse weather has been acknowledged, no existing benchmark provides a \textit{parametric}, severity controlled fog evaluation framework specifically targeting ground-to-air anti-UAV detection, where fog intensity is varied continuously as a function of a single scattering coefficient. This research addresses this gap by applying a physically inspired atmospheric scattering model to generate controlled fog severity levels \cite{agarwal_warlearn_2024}, enabling systematic robustness evaluation of anti-UAV detectors under adverse visibility conditions.


Finally, YOLOv5m is selected as the detection architecture, following Munir et al. (2024) \cite{munir_impact_2024} who demonstrate it outperforms later YOLO generations under adverse weather conditions. Using the same architecture and variant ensures that degradation results reported here are directly comparable to the closest related benchmark. A more in-depth research design can be found in the methodology section.

The study further dedicates a focused sub-analysis to UAV targets across size categories, with particular attention to small targets (those below 50 pixels, including the rare tiny instances below 10 pixels \cite{jiang_anti-uav_2023, huang_anti-uav410_2024}), which are especially vulnerable to visibility degradation due to their minimal pixel footprint and heightened sensitivity to resolution downscaling \cite{song_review_2025, hu_improved_2024, sdraka_deep_2022, wang_research_2026}. This study examines whether fog affects UAVs differently depending on their apparent image size, since distant or visually small UAVs may provide fewer visual cues for detection.

This study investigates the robustness of YOLOv5m-based UAV detection under
controlled fog degradation and evaluates whether synthetic fog augmentation
can improve performance under adverse visibility conditions without imposing
a substantial cost under clear-sky conditions. The analysis characterises
detector performance across ten progressively increasing fog severity levels,
with particular attention to changes in mean Average Precision ($mAP$),
precision, recall, and F1 score.

To examine whether visibility degradation affects targets differently
according to their apparent size, detection performance is further analysed
across small, medium, and large UAV categories. The clear-trained baseline is
then compared with a fog-aware model trained on a mixture of clear and
synthetically degraded images, allowing the robustness gains under fog to be
assessed alongside any corresponding reduction in clear-sky performance.

Through this controlled evaluation, the study establishes a
severity-dependent benchmark for ground-to-air anti-UAV detection under
synthetic fog. Rather than treating adverse visibility as a binary
clear-versus-fog condition, the proposed framework quantifies how detector
reliability changes across a continuous range of fog intensities and identifies
the visibility regimes in which performance degradation becomes most
pronounced.

\section{Related Work}
\label{sec:related_work}
Vision-based UAV detection has seen substantial progress, yet a persistent reliability gap remains: most systems are developed and evaluated under clear atmospheric conditions, while real-world deployments must operate under uncontrolled environmental degradation, particularly fog \cite{munir_impact_2024, ding_vision-based_2025}.
\subsection{Anti-UAV Tracking Benchmarks}
Large-scale benchmarks for vision-based object detection of UAVs have been designed and researched substantially. Jiang et al. \cite{jiang_anti-uav_2023} introduce Anti-UAV300, the first large-scale benchmark specifically designed for ground-to-air UAV tracking. Comprising of 318 RGB-T video pairs and 558k manually annotated bounding boxes, across diverse scenes and lighting condition, the benchmarks addresses challenges such as scale variation, occlusion, fast motion, and thermal crossover, and proposes a dual-flow semantic consistency (DFSC) training strategy that improves tracker robustness by leveraging cross-sequence semantic information. However, all sequences are recorded under clear atmospheric conditions, even on night time videos, leaving the benchmark's utility for evaluating detector robustness under adverse weather unaddressed.

Huang et al. \cite{huang_anti-uav410_2024} expanded this research by incorporating more realistic wild scenes with diverse backgrounds such as forests, mountains, and urban areas. Anti-UAV410 is a large-scale thermal infrared benchmark comprised of 410 sequences with over 438k manually annotated bounding boxes. Unlike Anti-UAV300 which features predominantly large-sized targets against simple sky or building backgrounds, Anti-UAV410 emphasizes tiny and small UAV targets, defining tiny targets as bounding boxes with a diagonal length below 10 pixels, along with challenging attributes such as dynamic background clutter and thermal crossover. However, like its predecessor, all sequences are captured under clear atmospheric conditions, providing no evaluation coverage for weather-induced visibility degradation.

Xie et al.\cite{xie_cst_2025} introduced CST Anti-UAV, another example of a thermal infrared benchmark of 220 sequences with over 240k annotated bounding boxes,  incorporating a substantially larger proportion of tiny UAV targets, 4.5 times more than existing datasets, alongside genuinely complex urban scenes with dynamic distractors such as pedestrians and vehicles. The benchmark is also the first to provide complete manual frame-level attribute annotations across all sequences, enabling more precise evaluation of tracker performance under specific challenges. Evaluation results reveal that even the state-of-the-art tracker achieves only $35.92\%$ state accuracy on CST Anti-UAV, compared to $67.69\%$ on Anti-UAV410, quantifying the performance gap that complex scenes and tiny targets introduce. However, CST Anti-UAV contains no sequences captured under adverse atmospheric conditions, leaving adverse weather unaddressed. 

Finally, Feng et al. \cite{feng_hazydet_2025} introduced HazyDet, the first large-scale benchmark specifically designed for drone-view object detection in hazy conditions, comprising 383k annotated instances drawn from both real-world foggy captures and physics-based synthetic scenes generated using the Atmospheric Scattering Model. To address haze-induced feature degradation, they propose DeCoDet, a depth-conditioned detector that dynamically modulates feature representations using estimated depth cues and a Progressive Domain Fine-Tuning strategy to bridge the synthetic-to-real domain gap, achieving $+1.5\%$ mAP over the closest competitor on real-world hazy data. While HazyDet represents an important step toward adverse-weather drone detection, it focuses on object detection from a UAV-to-ground perspective and does not address the ground-to-air anti-UAV tracking problem under fog, leaving the specific impact of fog severity on UAV target detection uncharacterised.

Recent multimodal anti-UAV studies have also examined reliability from the perspective of sensor uncertainty and fusion rather than atmospheric visibility. Suttorp et al.~\cite{suttorp_uncertainty_2026} evaluate evidential RGB--thermal fusion with conflict-aware belief aggregation, while Golovchits et al.~\cite{golovchits_evidential_2026} study evidential learning, Dempster--Shafer fusion, and uncertainty-driven sensor gating across thermal, RGB, audio, and RF anti-UAV settings. These studies broaden the analysis of anti-UAV reliability under modality disagreement and sensor-specific failure, but they do not provide a severity-controlled evaluation of fog-induced visibility degradation.
Beyond UAV-specific benchmarks, recent work on challenging
multi-object tracking settings has examined whether stronger visual
representations improve the detection and tracking of small or partially
occluded targets. Anwar and Mohammadi Ziabari~\cite{anwar2025attention}
compared CNN- and Transformer-based backbones within FairMOT and found that,
although Vision Transformer features are promising for small and occluded
objects, they do not consistently outperform the conventional CNN-based
backbone. Faber et al.~\cite{faber2024leveraging} similarly investigated
the transfer of DINOv2 foundation-model representations to FairMOT through
knowledge distillation. Their results showed improvements in selected
settings but no consistent advantage over the original FairMOT model.
Together, these findings indicate that stronger pretrained representations
alone do not necessarily resolve the challenges associated with difficult
visual targets. Neither study, however, evaluates robustness under
weather-induced visibility degradation.
Related work on difficult video-analysis settings has also explored
task-specific mechanisms for extracting informative visual evidence from
sparse events. Katona et al.~\cite{katona2025marine} proposed MARINE, which
combines motion-based frame selection with DINOv2 feature extraction for
detecting rare predator--prey interactions in animal videos. The study
demonstrates that pretrained visual representations can be combined with
task-specific frame-selection strategies when informative observations are
sparse. Although the application differs substantially from anti-UAV
detection, it further motivates evaluating representation and
data-selection strategies under conditions in which reliable visual
evidence is limited.
Across all major anti-UAV benchmarks, adverse atmospheric conditions remain systematically absent, and the only drone-view haze dataset operates from an incompatible perspective, leaving severity-controlled fog evaluation for ground-to-air UAV detection as an open and unaddressed problem.

\subsection{Weather Degradation and Robustness in Object Detection}
Munir et al. \cite{munir_impact_2024} provides a systematic evaluation of adverse weather effects on vision-based UAV detection, benchmarking YOLOv5, YOLOv8, Faster-RCNN, RetinaNet, and YOLO-NAS against rain, noise, and motion blur conditions. Their results reveal severe performance degradation, with the best-performing model losing up to $50.62$ mAP points under torrential rain and $77.0$ points under high motion blur, while a weather augmentation training strategy recovered substantial performance, YOLOv5 gained $+35.4$ mAP points in severe rain. However, their evaluation treats weather as a categorical perturbation rather than a continuous severity gradient, and does not isolate fog specifically or evaluate performance across controlled visibility levels.

Agarwal et al. \cite{agarwal_warlearn_2024} propose WARLearn, a weather-adaptive representation learning framework that fine-tunes the feature extractor of an existing detector on synthetically degraded images using Barlow Twins loss to align adverse weather \cite{zbontar_barlow_2021} features with their clear weather counterparts, requiring no additional parameters or inference overhead. Fog is simulated using the Atmospheric Scattering Model with an adjustable scattering coefficient $\beta$ varied across 10 discrete levels, enabling multi-severity evaluation, achieving $52.6\%$ mAP on the unseen real-world RTTS foggy dataset, surpassing prior SOTA by $2.1\%$. However, WARLearn is evaluated on general object detection datasets (PascalVOC, RTTS) rather than UAV-specific anti-drone scenarios, and does not analyze how detection performance degrades as a continuous function of fog severity for aerial targets.

\subsection{Hypothesis and Research Gap}
Based on the deficiencies identified in the sources, this research tests the following hypotheses:
\begin{enumerate}
    \item \textbf{H1:}  Increasing fog severity will progressively reduce detection performance, with the degradation distributed unevenly across severity levels.
    \item \textbf{H2:} The performance degradation caused by fog will be disproportionally greater for small UAV targets (bounding box size $< 50$ pixels) than for larger targets.
    \item \textbf{H3:} Training with synthetic fog augmentation will improve $mAP$ under fog conditions compared to a model trained only on clear images, without completely eliminating performance degradation.
    \item \textbf{H4:} The clear-sky performance cost of fog-aware training will be small relative to the robustness it provides under fog. 
\end{enumerate}

In conclusion, while literature has begun to acknowledge  the sensitivity of anti-UAV detection systems to adverse weather, it continues to treat adverse weather as a categorical perturbation rather than a continuous, severity-dependent variable. A significant gap persists in the systematic, parametric evaluation of fog severity effects on ground-to-air anti-UAV detection.

\section{Methodology}
\label{sec:methodology}

\subsection{Dataset}
\label{dataset}
This study uses the Anti-UAV300 dataset introduced by Jiang et al. \cite{jiang_anti-uav_2023}. The dataset comprises $318$ video sequences with over $500{,}000$ manually annotated bounding boxes across both RGB and infrared modalities, recorded under two lighting conditions (day and night) across diverse background scenes including buildings, clouds, and trees, using RGB and infrared cameras mounted on a rotating turret. This study exclusively uses the RGB modality, yielding $280{,}067$ annotated bounding boxes across $160$ training, $67$ validation, and $91$ test sequences, with extracted frames at a resolution of $1920 \times 1080$ pixels. The thermal stream is excluded to maintain evaluation of visible-light detection robustness under foggy conditions. The dataset is provided with predefined splits by Jiang et al. \cite{jiang_anti-uav_2023}: the training and validation sets derive from non-overlapping clips of the same videos, while the test set is drawn from entirely separate videos, ensuring an independent evaluation set. The scale distribution of the RGB modality exhibits a bimodal pattern with a median scale of $86$ pixels, reflecting two predominant UAV distance ranges. The dataset paper reports an average target scale below $40$ pixels across splits \cite{jiang_anti-uav_2023}; the higher median measured here for the RGB modality likely reflects that this reported average aggregates both modalities, with the lower resolution infrared stream reducing it. All scale statistics in this work are computed directly from the RGB annotations as $s=\sqrt{w\times h}$, following the dataset's convention \cite{jiang_anti-uav_2023}.

\begin{figure}[H]
    \centering
    \includegraphics[width=1\linewidth]{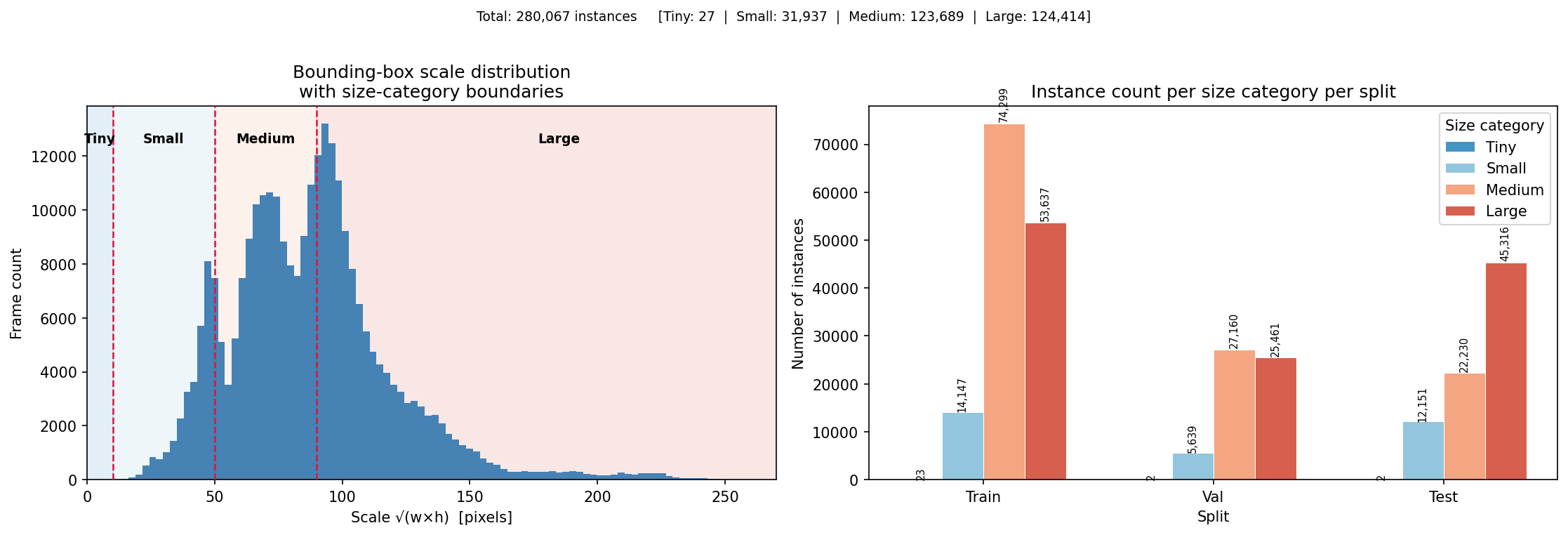}
    \caption{Bounding box scale distribution and size category breakdown per split for the RGB modality of the Anti-UAV300 dataset.}
    \label{fig:size_cat}
\end{figure}

\begin{figure}[htb]
    \centering
    \includegraphics[width=1\linewidth]{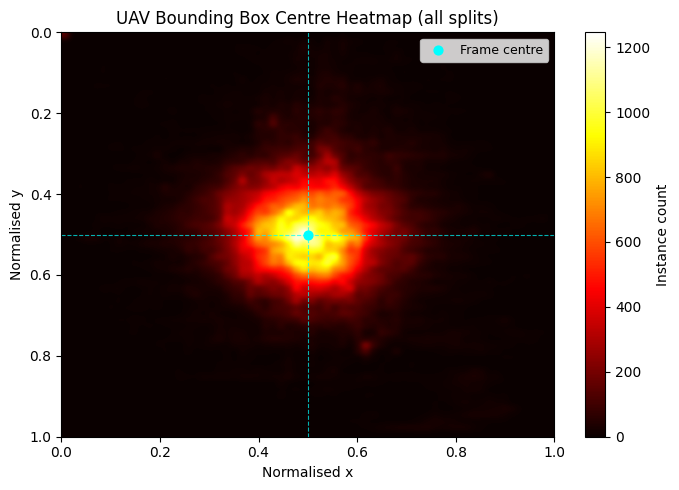}
    \caption{Spatial distribution of UAV bounding box centers across all splits.}
    \label{fig:heatmap}
\end{figure}

Notably, the test set exhibits a higher median scale ($93$ pixels) compared to the training set ($80$ pixels), meaning the baseline model is evaluated on slightly larger targets than it was trained on. UAV targets are stratified by scale to enable size dependent analysis. The anti-UAV literature size-categorizes targets into tiny, small, medium, and larger groups \cite{xie_cst_2025, huang_anti-uav410_2024}, adopting a tiny threshold of $10$ px to mark the reliable detection limit. These schemes are developed for thermal-infrared datasets dominated by very small targets, however, their upper boundaries are unsuited to the RGB modality studied here, with substantially larger targets. As no prior work provides a size taxonomy for the RGB modality of Anti-UAV300, category boundaries are defined directly from its measured scale distribution (Figure \ref{fig:size_cat}): tiny ($s<10$px), small ($s<50$px), medium ($s<90$px), and large ($s\ge90$px). Standard COCO thresholds are unsuitable, being calibrated for large natural image objects, they would assign nearly all UAV targets to a single category and provide no scale discrimination \cite{noauthor_benchmarking_nodate, zhang_learning_2026}. Of the $280{,}067$ instances, only 27 fall in the tiny category (and only 2 in the test split). Thus, they are merged into the small category for the size-stratified evaluation, as too few yield reliable per category metrics. Merging the tiny instances into the small category prevents the reporting of a misleading precise tiny-category metric, but it does not recover statistical power. Any tiny-specific behavior is averaged into the larger small band rather than resolved. Thus, findings at the small scale remain provisional. 
Figure \ref{fig:heatmap} shows that UAV targets are predominantly concentrated around the image center across all splits, placing most targets in the region where the radial fog prior applies its strongest attenuation. As Section \ref{sec:prior} establishes, this central concentration is an artifact of operator turret-tracking rather than a true depth relationship. 

Since fog is synthetically applied to the clear RGB frames post-hoc, the original manually annotated bounding boxes remain valid as ground truth across all fog severity levels, requiring no re-annotation.

\subsection{Research Design}
This study follows a controlled experimental design in which fog  severity is the single manipulated variable. Synthetic fog is applied post-hoc to clear test images at ten graded intensities so that the scene content, target positions, and ground-truth annotations remain identical across all conditions. The only variation is visibility, which allows to isolate the effect of fog detection accuracy from confounding scene differences. Two models: \textit{clear-trained baseline} and \textit{fog-aware model} trained on mixed clear and foggy data evaluated on the held-out test set at each severity level, and their performance is compared. Because the architecture, training schedule and hyperparameters are held fixed across both models, observed differences in degradation are attributable to training data composition rather than model capacity. The overall design is illustrated in Appendix \ref{sec:apx:methodology_vis}.

\subsection{Fog Generation Method}
\label{sec:fog_generation}
Synthetic fog is applied to the clear dataset using the physically-based atmospheric scattering model of Agarwal et al. \cite{agarwal_warlearn_2024}. Ten fog severity levels are generated by varying the atmospheric scattering coefficient according to $\beta=0.01i+0.05$ for $i\in\{0,1,\ldots,9\}$, with atmospheric light fixed at $A=0.5$. The complete formulation and depth approximation strategy are detailed in Appendix \ref{sec:apx:fog_generation}.

It is important to emphasize that this formulation should not be interpreted as a physically accurate depth map. Rather, it functions as a \textit{heuristic degradation protocol}: a spatially structured, deterministic prior that applies stronger fog attenuation toward the image center and weaker attenuation near the periphery. The prior is fully deterministic and resolution-dependent, ensuring reproducible fog generation across the dataset without requiring depth estimation. The validity of treating this prior as a heuristic rather than a physical depth is examined in Section \ref{sec:prior}.

\begin{figure*}[t]
  \centering
  \includegraphics[width=\linewidth]{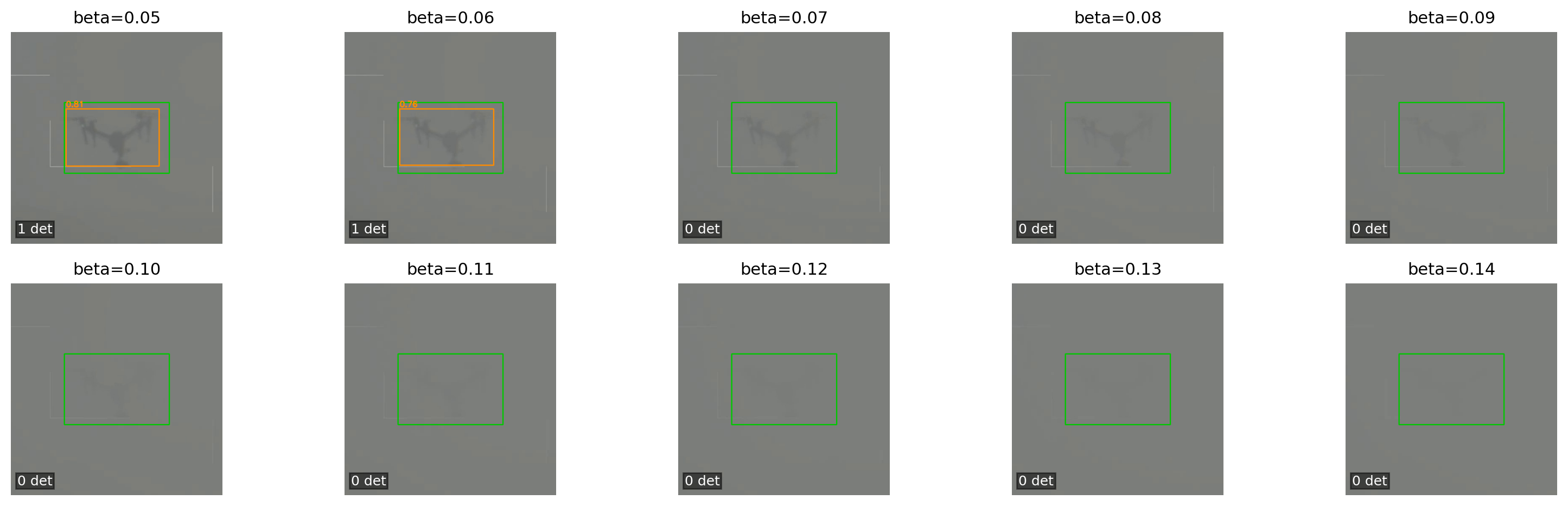}
  \caption{Baseline YOLOv5m detections on one Anti-UAV300 frame across all ten fog levels ($\beta$ above each panel). Orange: predictions with confidence; green: ground truth. The target is detected only at $\beta = 0.05$--$0.06$  with falling confidence, then lost from $\beta = 0.07$ on, though it remains present throughout.}
  \label{fig:fog_degradation}
\end{figure*}

To avoid data leakage, each clear video and all of its fog-generated variants are kept within the same dataset split (training, validation, or test).

\subsection{Models}
\label{models}
To quantify the impact of fog on drone detection, two YOLOv5m variants are trained: a clear-sky baseline and a fog-aware model.

YOLOv5m is selected as the detection architecture, following Munir et al. \cite{munir_impact_2024} who demonstrate it outperforms later YOLO generations under adverse weather conditions. With approximately $21.2M$ parameters \cite{ultralytics_ultralytics_nodate}, YOLOv5m offers a strong balance between detection accuracy and inference speed, and its performance under adverse weather has been directly benchmarked in the most closely related work, ensuring that degradation results reported here are directly comparable \cite{munir_impact_2024}. Fixing a single architecture as a controlled variable ensures that measured changes in $mAP$ are attributable solely to fog severity rather than architectural differences; cross-architecture comparison is left as future work. Both variants are initialized from YOLOv5m weights pretrained on the COCO dataset \cite{ultralytics_ultralytics_nodate} and fine-tuned on the Anti-UAV300 training split, the standard transfer-learning setting for object detection. Since both models start from the same pretrained checkpoint and differ only in their training-data composition, the initialization is a shared constant and does not confound the comparison between them. Both models are trained for 300 epochs with early stopping (patience 100) triggered based on validation mAP@0.5:0.95. The variants only differ in their training and validation data:
\begin{enumerate}
    \item \textbf{Clear-sky baseline:} trained exclusively on clear-weather data, with early stopping on the clear-only validation set. This variant establishes reference performance under ideal conditions.
    \item \textbf{Fog-aware variant:} trained on a $50{:}50$ clear/fog-augmented  training set with matching validation distribution. All other hyperparameters remain unchanged. 
\end{enumerate}

Unless otherwise stated, all remaining hyperparameters follow the Ultralytics YOLOv5 defaults, including an initial learning rate of 0.01, and an input resolution of $640\times640$ pixels \cite{ultralytics_ultralytics_nodate}. The batch size was set to 32 (default 16) to make better use of the A100's 32\,GB of memory. No other hyperparameters were tuned.
All experiments are conducted on the Snellius national supercomputer using a single NVIDIA A100 GPU with 32GB memory, ensuring reproducibility of reported results.

\subsection{Radial Prior}
\label{sec:prior}
To empirically assess whether the radial distance prior captures meaningful depth structure in the dataset, the relationship between bounding box scale and radial distance from the image center is analyzed across all $280{,}067$ annotated instances. A very weak negative correlation was observed, Pearson $r=-0.085$ and Spearman $\rho = -0.0934$, which explains less than $1\%$ of variance ($r^2=0.007$). This weak relationship demonstrates that the prior does not capture true scene depth; it is instead a deterministic, spatially-structured heuristic degradation protocol. 

\begin{figure}[H]
    \centering
    \includegraphics[width=0.9\linewidth]{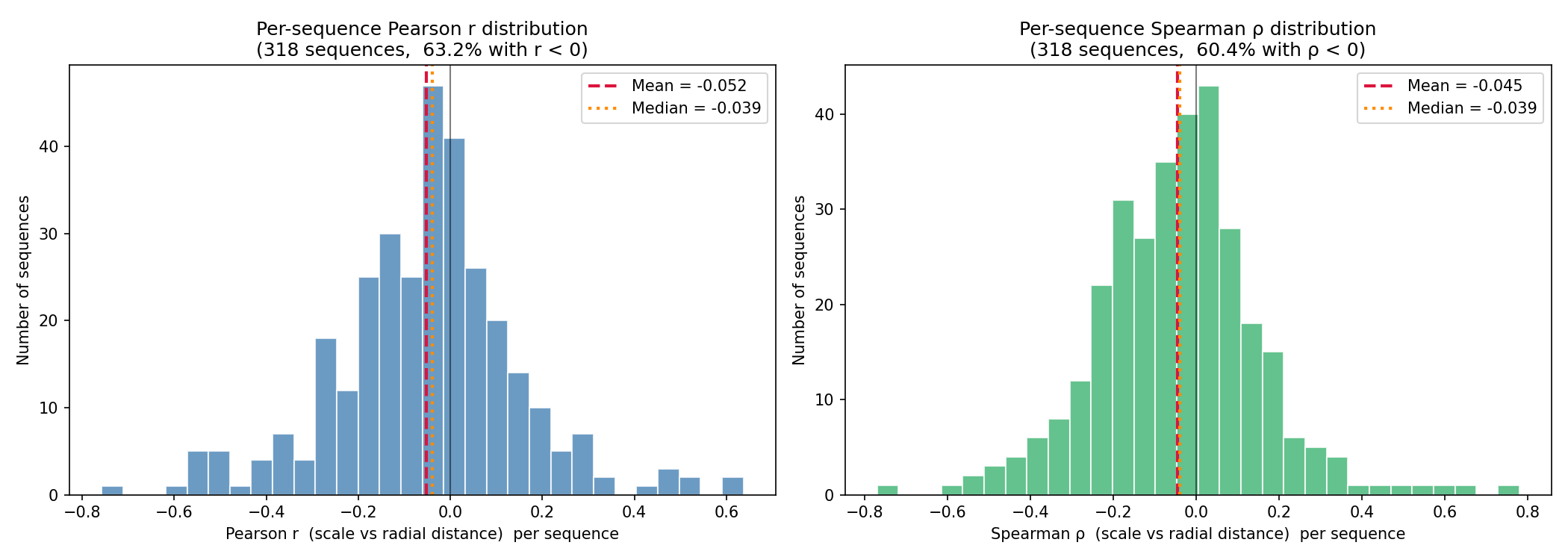}
    \caption{Per-sequence Pearson and Spearman correlation between bounding-box scale and radial distance from image center (318 sequences).}
    \label{fig:radial_within_seq}
\end{figure}

Per-sequence analysis confirms the relationship is highly variable across sequences (mean $r=-0.052$, $63.2\%$ of sequences show negative $r$, Figure \ref{fig:radial_within_seq}), which is expected given that UAV targets appear near the image center primarily because operators actively track and center the targets using a rotating turret, rather than due to a systematic geometric depth relationship.
This variability and weakness of correlation confirm that the prior is fundamentally a reproducible degradation protocol, a mathematical tool that applies spatially-structured fog attenuation without capturing physical depth relationships. Its utility lies in deterministic, dataset-agnostic fog generation, not in depth fidelity.

\begin{figure}
    \centering
    \includegraphics[width=1\linewidth]{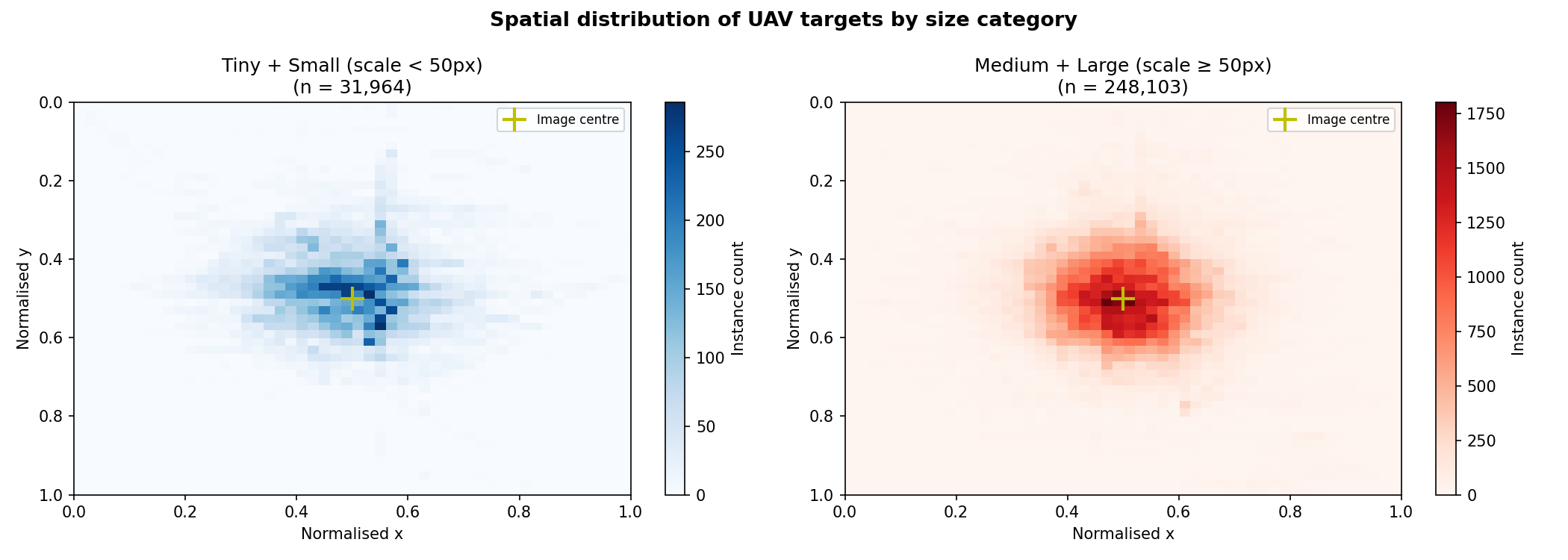}
    \caption{UAV box centers by size category. Large and medium targets  cluster more centrally where radial prior attenuates fog the most; small and tiny targets also mostly cluster centrally, with some outliers.}
    \label{fig:radial_heatmap}
\end{figure}

Stratified spatial analysis further reveals that the tiny and small targets ($ <50 \,px$, $n=31{,}964$) concentrate around the image center, as do the medium and large targets ($\geq50\,px$, $n=248{,}103$), which cluster even more tightly there. Because the radial prior applies strongest attenuation toward the image center, the bulk of targets does not preferentially penalize small targets as they are predominantly central, not peripheral. In reality, the most concentrated central mass belongs to larger targets, so the prior's heaviest attenuation falls more on the large central majority. Opposite of a physically realistic model, in which distant, small targets would be more obscured. (Figure \ref{fig:radial_heatmap}).

\subsection{Evaluation Protocol}
\label{evaluation_metrics}
To evaluate the impact of fog, each clear test video has corresponding fog-augmented versions generated using the atmospheric scattering model described in Section \ref{sec:fog_generation}. All 10 discrete fog severity levels ($i \in \{0,1,\ldots,9\}$) are evaluated to enable a continuous severity-degradation analysis, directly supporting the novelty claim of severity-controlled robustness evaluation.

Because fog is synthetically generated from annotated clear images, the original manually labeled bounding boxes remain valid and serve as ground truth for all fog levels \cite{jiang_anti-uav_2023}.

\paragraph{Primary Metrics}
Detection performance is evaluated on a frame-wise basis using standard object detection metrics. Since this study focuses on detector robustness rather than temporal tracking, video-level consistency is not explicitly modeled. 

\begin{itemize}
    \item Mean Average Precision ($mAP@0.5:0.95$)
    \item Mean Average Precision ($mAP@0.5$)
    \item Precision
    \item Recall
    \item F1 Score
\end{itemize}

Evaluation is conducted separately for:
\begin{enumerate}
    \item Clear test set (baseline performance)
    \item Fog test set, per severity level ($i \in \{0,\ldots,9\}$)
\end{enumerate}

Performance degradation due to fog is quantified as: 
\begin{equation}
    \Delta mAP=mAP_{clear} - mAP_{fog}
\end{equation}
This measures robustness loss across all ten severity levels.

\paragraph{Size-Stratified Evaluation}
Since drone detection is particularly sensitive to small-object visibility degradation \cite{xie_cst_2025, jiang_anti-uav_2023, huang_anti-uav410_2024}, results are analyzed by object size category defined by geometric mean scale $s = \sqrt{w \times h}$: Small $[0, 50)$,  Medium $[50, 90)$, Large $[90, \infty)$, adapted to the RGB modality resolution of the Anti-UAV300 dataset \cite{jiang_anti-uav_2023}. 

\paragraph{Failure Mechanism Analysis}
To characterize failure mode beyond aggregate performance, the following additional analyses are conducted:
\begin{itemize}
    \item Precision-recall curves per fog severity level, to characterize how the precision-recall trade-off evolves as fog density increases.
    \item Confidence score distributions under clear  vs. fog conditions, to distinguish missed detections from low-confidence suppression.
    \item False positive rate analysis across fog levels, to assess whether fog introduces spurious detections.
    \item IoU stability analysis, to assess whether surviving detections at high fog severity are 
    well-localized or merely above the confidence threshold.
\end{itemize}

\paragraph{Secondary Analyses}
Results are additionally stratified by:
\begin{itemize}
    \item \textbf{Input resolution} 
    ($640 \times 640$ vs. $1280 \times 1280$): 
    downscaling may cause small UAV targets to lose discriminative detail before inference. Thus, to assess fog-induced small-target degradation reflects genuine atmospheric visibility loss or is partly an artifact of this resolution reduction, inference is additionally conducted at  $1280 \times 1280$ on small-targets at the baseline model. 
    \item \textbf{Atmospheric light sensitivity} 
    ($A \in \{0.3, 0.5, 0.7\}$): evaluations at three atmospheric light values assess the stability of conclusions under different fog parameterizations, directly addressing the sensitivity concern raised regarding the fixed $A = 0.5$ choice. The light sensitivity is evaluated on three representative fog severity levels: $i\in \{2,5,9\}$, representing low, medium and high fog. 
    \item \textbf{Uniform-fog control}: to assess whether the degradation curve reflects fog density or the spatial geometry of the radial prior, the baseline model is additionally evaluated on spatially uniform fog at the same ten $\beta$ levels, with transmission depth held constant at the mean of the radial depth map per level. If the radial prior's geometry were driving the collapse, the uniform curve should differ materially in shape from the radial baseline.
\end{itemize}

Secondary analyses are conducted on a single representative run per model (baseline seed 1, fog-aware seed 1) and reported without seed aggregation. These serve to characterize failure mechanisms and parameter sensitivity rather than to measure core degradation, and are interpreted descriptively rather than as statistically quantified claims.
\paragraph{Statistical Reliability}
\label{sec:replicability}
To quantify training stochasticity, the two primary models, the clear-sky baseline and the fog-aware variant, are each trained three times with independent random seeds (seeds 1, 2, 3). All primary-model metrics, across clear and all ten fog severity levels ($i = 0,\ldots,9$), are reported as mean $\pm$ standard deviation across the three runs; this spread reflects sensitivity to random initialization.

\section{Results}
\label{sec:results}

\subsection{Degradation Magnitude}
\begin{figure}[t]
    \centering
    \includegraphics[width=1\linewidth]{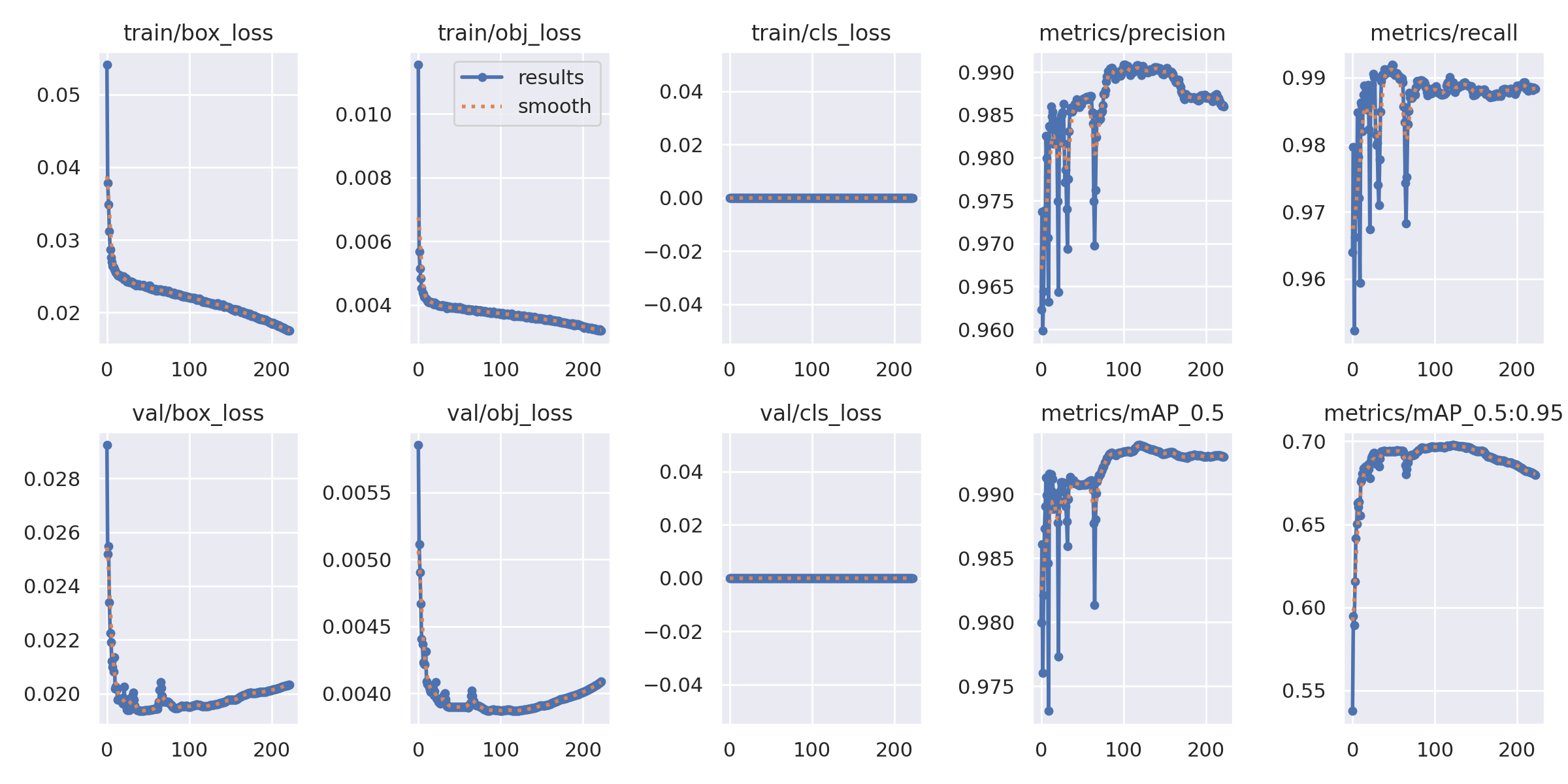}
    \caption{Training and validation curves for the baseline YOLOv5m model trained  on clear sky data. Training halted at epoch 223. Best epoch 123, validation mAP@0.5:0.95 $=0.698$ is used for all inference. Validation clips derive from the same source videos as training, all reported performance uses the held-out test set.}
    \label{fig:baseline_results}
\end{figure}

Figure \ref{fig:baseline_results} confirms that the baseline YOLOv5m model (seed 1) converges on clear-sky data, with the best checkpoint at epoch 123 (validation mAP@0.5:0.95 of $0.698$). All reported test-set results below are aggregated across three independent training runs and reported as mean $\pm$ std (Section \ref{sec:replicability}). See Figure \ref{fig:fog_degradation} for visual representation of the baseline's detection.

On the held-out test set under clear conditions, the baseline reaches $0.567 \pm 0.002$ mAP@0.5:0.95 ($0.956 \pm 0.001$ mAP@0.5) across the three training seeds, while the fog-aware model reaches $0.515 \pm 0.006$ ($0.938 \pm 0.009$ mAP@0.5); full per-metric results are reported in Table \ref{tab:clear_tradeoff}. The small standard deviations indicate stable convergence across random seeds. 

These metrics confirm robust clear sky detection on the Anti-UAV300 RGB modality. The small standard deviations indicate stable convergence across random seeds.

\subsection{Fog-Aware Recovery Under Fog Training}
The fog-aware model, trained on a $50{:}50$ mixture of clear and fog-augmented images across all ten severity levels, substantially outperforms the baseline under all fog conditions. The largest absolute gain occurs at moderate fog ($+0.320$ mAP@0.5:0.95 at $i=4$), with comparable improvements across the moderate range ($i=3$: $+0.314$; $i=5$: $+0.299$). At heavy fog ($i=9$) the fog-aware model achieves $0.093 \pm 0.002$ compared to the baseline's $0.022 \pm0.003$ ($+0.071$). Recovery is therefore broad across the moderate band rather than concentrated at a single severity. However, it does not eliminate the collapse, the model still falls from $0.515$ under clear conditions to $0.093$ at $i=9$.

\begin{figure}[t]
\centering
\includegraphics[width=1\linewidth]{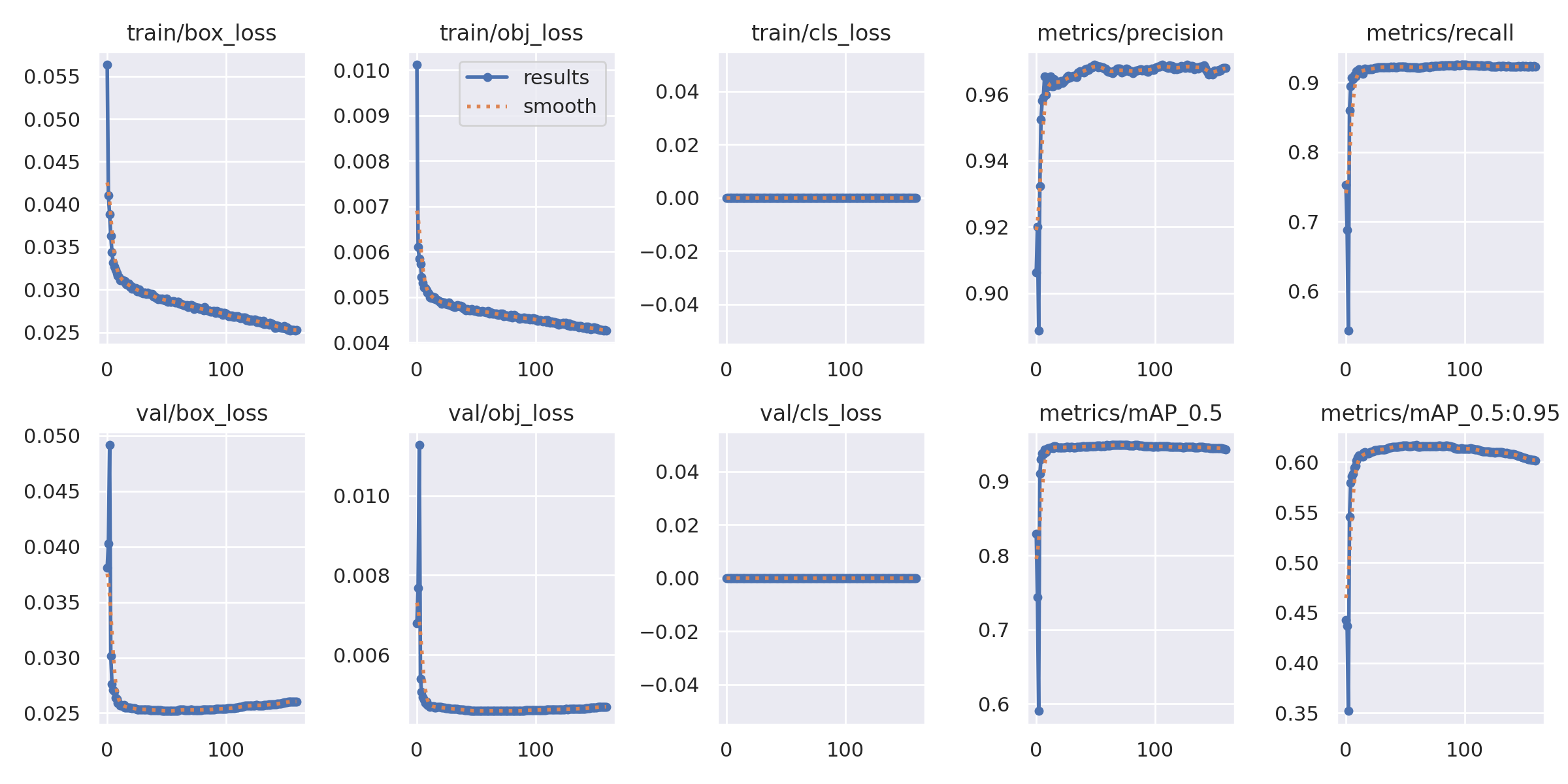}
\caption{Training and validation curves for the fog-aware YOLOv5m model, trained on a $50{:}50$ mix of clear and fog-augmented data ($i \in \{0,\ldots,9\}$, sampled uniformly) for up to 300 epochs with early stopping (patience 100); training halted at epoch 160. Validation mAP@0.5:0.95 peaks at the best checkpoint (epoch $59$, $0.617$) before a slight decline that triggered early stopping.}
\label{fig:fogaware_results}
\end{figure}

Figure \ref{fig:fogaware_results} shows the training curves for the fog-aware model for a single-seed (seed 1) representative run. 

\subsection{Clear-Sky Cost of Mixed Training}
Under clear-sky conditions, the fog-aware model achieves mAP@0.5:0.95 of $0.515 \pm 0.006$, compared to $0.567 \pm0.002$ for the clear-trained baseline, a reduction of $0.052$ ($9.1\%$). Precision, recall and F1 show similarly small reductions (Table \ref{tab:clear_tradeoff}).

\begin{table}[t]
\centering
\begin{tabular}{lcc}
\toprule
\textbf{Metric} & \textbf{Baseline} & \textbf{Fog-aware} \\
\midrule
Precision & $0.971\pm0.003$ & $0.946\pm0.012$ \\
Recall & $0.919 \pm 0.003$ & $0.896\pm0.009$ \\
F1 & $0.944\pm0.001$ &$0.920\pm0.009$\\
mAP@0.5 & $0.956\pm0.001$ & $0.938 \pm0.009$\\
mAP0.5:0.95 & $0.567\pm0.002$ & $0.515\pm0.006$ \\
\bottomrule
\end{tabular}
\caption{Clear-sky test performance of the baseline and fog-aware models (mean $\pm$ std across three seeds). Mixed training reduces clear-sky mAP@0.5:0.95 by $0.052$ ($9.1\%$).}
\label{tab:clear_tradeoff}
\end{table}

Figure \ref{fig:degradation} shows mAP@0.5:0.95 and recall for both models across clear and all ten fog severities. The baseline falls steeply from clear through $i=2$, then flattens toward a near-zero floor by $i=5$. The fog-aware model stays above the baseline at every level.

Figure \ref{fig:delta_map} illustrates the loss relative to each model's clear-sky baseline. The baseline loses $0.194$ mAP@0.5:0.95 at the lightest fog ($i=0$) and $0.545$ at the heaviest ($i=9$). The fog-aware model shows smaller degradation at low fog levels, though losses accumulate at higher severities.

To test whether the radial prior's spatial geometry drives the degradation curve, the baseline was additionally evaluated on spatially uniform fog at the same ten $\beta$ levels (transmission depth held constant at the mean radial depth per level; single representative run). The uniform-fog baseline degrades from $0.492$ at $i=0$ to $0.114$ at $i=9$, following the same non-linear shape as the radial curve but at consistently higher mAP at every level (Figure~\ref{fig:degradation}).

\subsection{Failure Mechanisms}
\label{sec:failure_mech}
The confidence distributions (Figure \ref{fig:conf_dist}) show that under fog the baseline loses confident true positives entirely from $i\approx4$, whereas the fog-aware model retains a peak near $0.8$ at every severity. The false-discovery rate (Figure \ref{fig:fp_rate}) rises monotonically for both but stays lower for the fog-aware model from $i=2$ onward. However, localization is preserved: the mean IoU of surviving the true positives holds at $\sim0.75$ across all fog levels for both models (Figure \ref{fig:iou_stab}). Fog therefore degrades detection by suppressing recall and confidence rather than by displacing the boxes that survive. Per severity precision-recall curves are reported in Figure \ref{fig:pr_curves}.

\subsection{Fog Severity Across Drone Size Categories}

Because the RGB modality contains only 27 tiny instances (merged into small; Section \ref{dataset}), small-category results are read as scale-dependent trends rather than definitive conclusions. Under clear conditions the three groups lie within ${\sim}0.07$ mAP@0.5:0.95, large being strongest. Under fog, large targets stay most robust through $i\le7$ but fall below small at $i\ge8$ as all groups approach the floor; medium degrades fastest, dropping below small from $i=3$. The fog-aware model preserves the large${>}$medium${>}$small ordering at every severity.
 See the full table in Table \ref{tab:size_results}.

To test whether small-target collapse under fog reflects atmospheric visibility loss or resolution loss from downscaling $1920\times1080$ frames to $640\times640$, small-target inference was additionally run at $1280\times1280$ using the baseline model trained at $640\times640$. Because the model is trained at $640$ and evaluated at $1280$, this comparison carries a train-inference resolution mismatch and does not cleanly isolate resolution from fog. Small-target recall and mAP@0.5:0.95 at $1280$ are lower than at $640$ at every condition (e.g.\ clear recall $0.947\to0.917$; $i=2$ recall $0.624\to0.517$. See Table \ref{tab:resolution} for full results.

\subsubsection{Atmospheric light sensitivity analysis}
\begin{figure}[H]
    \centering
    \includegraphics[width=1\linewidth]{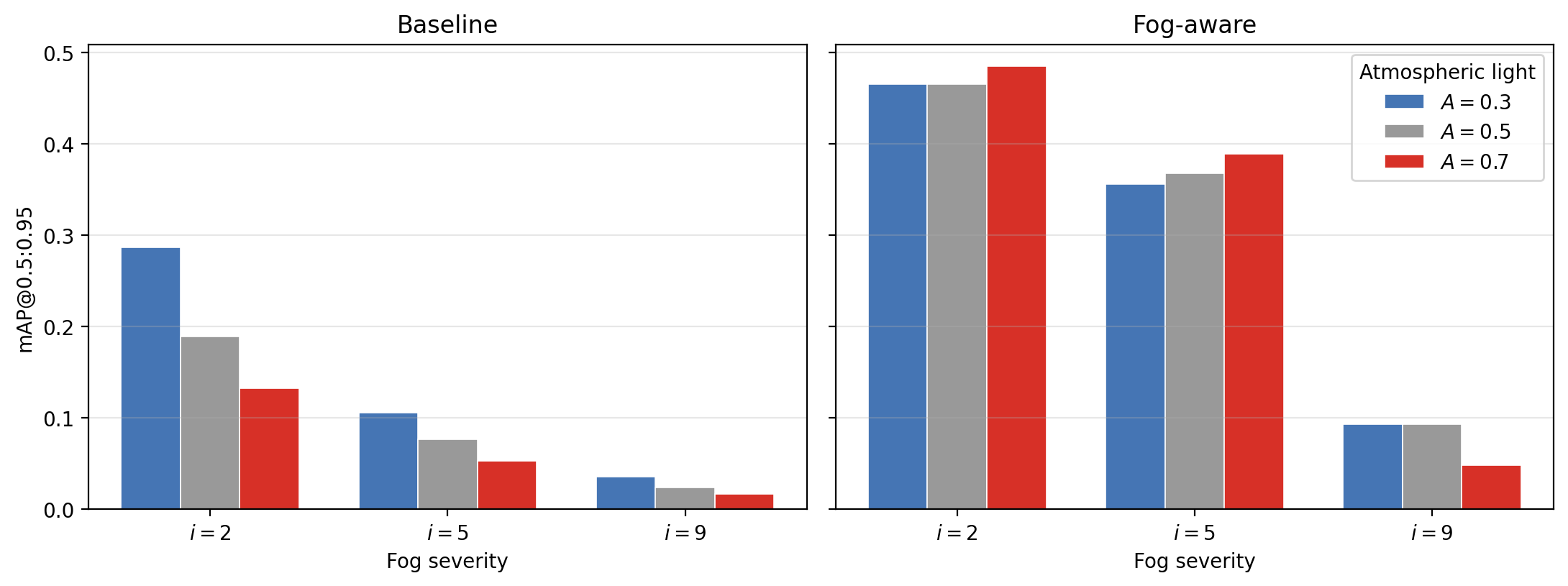}
    \caption{Atmospheric light sensitivity at fog levels $i \in \{2,5,9\}$ for $A \in \{0.3, 0.5, 0.7\}$ (mAP@0.5:0.95; single representative run per model).}    \label{fig:sensitivity}
\end{figure}

To assess whether the degradation findings depend on the atmospheric light parameter $A$, both models were evaluated at $A \in \{0.3, 0.5, 0.7\}$ at the three reference fog levels $i \in \{2, 5, 9\}$ (low, moderate, high), with $A=0.5$ as the primary condition (Section \ref{evaluation_metrics}). Since $A$ is a nuisance parameter rather than a primary axis, three representative severities suffice to test whether the conclusions hold across fog parameterizations. Figure \ref{fig:sensitivity} presents the results.

\subsection{Statistical Reliability}
\label{sec:statistical_reliability}
Both baseline and fog-aware models are trained three times with independent random seeds. All metrics are reported as mean $\pm$ standard deviation (std) across seeds, capturing training stochasticity due to random initialization and optimization dynamics. 

Table \ref{tab:full_degradation} reports mAP@0.5:0.95 across all eleven evaluation conditions: clear and all ten fog severity levels. The standard deviation is remarkably small across all conditions (std $\leq0.006$ for baseline, $\leq 0.007$ for fog-aware), indicating stable and reproducible training across random seeds.

\begin{table}[H]
\centering
\small
\begin{tabular}{lcc}
\toprule
     \textbf{Fog Severity}& \textbf{Baseline}& \textbf{Fog-aware}  \\
     \midrule
     Clear & $0.567 \pm 0.0017$&$0.515 \pm 0.0064$\\
     $\beta = 0.05$&$0.373 \pm 0.0168$&$0.496 \pm 0.0006$\\
     $\beta=0.06$ & $0.278 \pm 0.0204$ & $0.482 \pm 0.0020$ \\
     $\beta=0.07$ & $0.188 \pm 0.0165$ & $0.463 \pm 0.0038$ \\
     $\beta=0.08$ & $0.122 \pm 0.0081$ & $0.437 \pm 0.0049$ \\
    $\beta=0.09$ & $0.088 \pm 0.0090$ & $0.408 \pm 0.0056$ \\
    $\beta=0.10$ & $0.066 \pm 0.0101$ & $0.365 \pm 0.0030$ \\
    $\beta=0.11$ & $0.047 \pm 0.0057$ & $0.314 \pm 0.0035$ \\
    $\beta=0.12$ & $0.038 \pm 0.0062$ & $0.243 \pm 0.0020$ \\
    $\beta=0.13$ & $0.029 \pm 0.0053$ & $0.165 \pm 0.0025$ \\
    $\beta=0.14$ & $0.022 \pm 0.0025$ & $0.093 \pm 0.0017$ \\
    \bottomrule
\end{tabular}
\caption{Mean Average Precision (mAP@0.5:0.95) across three training seeds for all fog severity levels.}
\label{tab:full_degradation}
\end{table}

\FloatBarrier

\section{Discussion}
\label{sec:discussion}

\subsection{Comparison with Prior Work}
Munir et al. \cite{munir_impact_2024} benchmark the same YOLOv5m architecture used here, reporting results in mAP@0.5. This allows direct comparison with the findings. 
In Munir et al.'s torrential rain, YOLOv5m falls from $95.6$ to $47.9$ mAP@0.5, a $49.9\%$ reduction ($47.7$ points). The worst single condition, heavy motion blur, reduces $95.6$ to $21.9$ mAP@0.5, a $77.1\%$ decline. Here, the baseline falls from $0.956$ to $0.048$ mAP@0.5 at the heaviest fog, a $95\%$ reduction ($90.8$ points). Thus, fog at $\beta=0.14$ harms detection more than even the most extreme rain, and exceeds the most severe motion blur from prior work. This is consistent with dense fog degrading scene contrast more uniformly than rain streaks. 

The mitigation trends align equally well. Munir et al. recover $+35.4$ mAP@0.5 in torrential rain through weather augmentation \cite{munir_impact_2024}.  The fog-aware model here gains $+60.0$ mAP@0.5 at moderate fog ($i=4$, $0.229 \to 0.829$). 

Agarwal et al.'s WARLearn, also used in this paper, adapts YOLOv3 to fog via feature alignment and achieves $52.6\%$ mAP@0.5 on the RTTS foggy set, beating the next-best (DENet) by $2.1\%$ \cite{agarwal_warlearn_2024}. Unlike the sharp decline seen here, WARLearn reports stable mAP@0.5 across $\beta = 0.05-0.14$. This likely reflects task differences: general objects in busy scenes versus UAVs against uniform backgrounds, where fog erases the little contrast present. Both studies show synthetic fog helps, but the sharper drop in this study is specific to anti-UAV detection.

\subsection{Critical Visibility Threshold (H1)}
H1 predicted degradation across fog severity would be non-uniform, that intermediate severity performance could not be recovered by interpolating between clear and heavy fog. This is supported, and the failure-mechanism analysis further reveals the degradation arises from reduced confidence and recall rather than localization error.

Performance loss is front-loaded: the single steepest drop occurs at the onset of fog (clear $\rightarrow i=0$, $\beta=0.05$), which alone accounts for about a third of the total mAP@0.5:0.95 loss ($0.194$ of $0.545$): roughly $70\%$ of the total loss occurs by $i=2$ and over $90\%$ by $i=5$, after which the curves flattens near a low floor (the baseline retains only $0.066$ mAP@0.5:0.95 at $i=5$). The most significant operational impact occurs across light-to-moderate fog ($\beta \approx 0.05-0.10$), not at the extremes.

Failure analysis shows the cause is loss of recall and confidence: mean IoU for true positives stays near $\sim 0.75$ at all fog levels, indicating surviving detections remain well-localized. Fog reduces target-background contrast until detections fall below the confidence threshold and disappear. Recall drops from $0.919$ (clear) to $0.382$ at $i=2$, confirming the early loss is genuine, not a floor effect.
Uniform-fog controls confirm this pattern is due to fog density, not the radial prior's geometry. The same early, steep decline is seen under uniform fog (from $0.492$ at $i=0$ to $0.114$ at $i=9$), with the steepest losses at lower severities. The radial prior yields consistently lower performance because it concentrates fog where most targets are, making it a harder test.
This front-loaded loss is what continuous severity control reveals and categorical clear-vs-adverse evaluation misses: most reliability is lost within a narrow range of moderate fog, so two-point testing misjudges where and how quickly detection fails.

\subsection{Size-Dependent Degradation (H2)}

H2 predicted that fog would harm small targets ($<50$ px) disproportionately more than larger ones. The results do not support this. In the baseline, medium targets degrade fastest ($98.3\%$ relative loss, clear to $i=9$), followed by large ($97.5\%$) and small ($96.0\%$). In the fog-aware model, large targets lose $76.9\%$, medium $84.6\%$, and small $86.8\%$. In neither model do small targets degrade disproportionately more than large targets, contradicting H2. Since both small and large targets cluster predominantly at the image center (Figure \ref{fig:radial_heatmap}), the radial prior applies comparable fog attenuation to all size groups, and does not artificially create or suppress a size-dependent difference.

To assess whether the 640$\times$640 training resolution artificially suppresses small-target performance independent of fog, inference was additionally run at $1280\times1280$ using the baseline model. Small-target recall and mAP@0.5:0.95 at 1280 are \emph{lower} than at $640$ at every condition (Table \ref{tab:resolution}). Rather than recovering small-target performance, higher inference resolution degrades it further, suggesting the issue is not downscaling at test time but the model's representations learned at $640\times640$ training resolution. Since the model was never trained at $1280\times1280$, the train-inference mismatch prevents a clean interpretation. A definitive test requires retraining at native resolution, left for future work.

H2 is therefore not supported. The resolution confound does not rescue H2, as higher inference resolution does not improve small-target performance. However, because the underlying model representations were trained at low resolution, the result remains \textbf{inconclusive}. The results cannot fully separate fog-induced small-target degradation from resolution-induced sensitivity limitations.

\subsection{Effectiveness and Cost of Fog-Aware Training (H3, H4)}
H3, that fog-aware training improves performance under fog without fully eliminating degradation, is supported. The fog-aware model exceeds the baseline at every fog level, with the largest gain at moderate severity ($+0.320$ mAP@0.5:0.95 at $i=4$) and a remaining gain at heavy fog ($0.093$ vs.\ $0.022$ at $i=9$). But it does not eliminate degradation: the fog-aware model drops from $0.515$ under clear conditions to $0.093$ at $i=9$. Thus, H3 holds; mixed training delays collapse but does not remove the visibility threshold. 

H4, that the clear-sky cost of fog-aware training is small compared to the gain, is also supported. Clear-sky mAP@0.5:0.95 drops only from $0.567$ to $0.515$ ($0.052$, $9.1\%$), with similarly small drops in precision, recall, and F1 (Table \ref{tab:clear_tradeoff}). 
The clear-sky costs of fog-aware training align with prior fog augmentation research. Mai et al. \cite{mai_3d_2021}, apply a similar mixed clear/fog approach to a 3D LiDAR-camera detector on fogged KITTI data. Fog initially reduced performance by $42.67$ for \emph{}{Moderate} objects, but using mixed training strategy recovered $26.72$ of that loss while incurring only an $8.23\%$ clear-sky cost, demonstrating the same trade-off pattern observed in this work ($9.1\%$). Kumar and Chadha \cite{kumar_fog_2025} show that dehazing as preprocessing improves fog detection but can unexpectedly degrade clear-images performance by introducing artifacts and false positives. More broadly, trade-offs introduced by image-level data interventions have
also been observed outside adverse-weather detection. Hooftman
et al.~\cite{hooftman2023cyclegan} used CycleGAN-based data augmentation
for gender classification and found that the intervention reduced measured
bias while also decreasing classification accuracy. Although the objective
and application differ from the present study, the result illustrates a
general property of augmentation-based interventions: improving robustness
or another target criterion does not necessarily preserve performance on
the original data distribution. This mirrors the present results, where the clear-sky cost emerges as a higher false-discovery rate (Figure \ref{fig:fp_rate}) rather than missed detections. The acceptability of this trade-off depends on the deployment context; in critical airspace protection, prioritizing fog detection over clear-sky performance is an intentional decision. 

The failure-mechanism analysis sharpens this trade-off: the fog-aware model has a marginally higher false-discovery rate under clear conditions (Figure \ref{fig:fp_rate}), but suppresses false positives and retains confident true positives at every fog level (Figure \ref{fig:conf_dist}). The small clear-sky cost is the price of training on degraded images, outweighed by the robustness gained across the fog range.

The atmospheric-light sensitivity analysis serves primarily as a robustness check on the headline result: the fog-aware model outperforms the baseline at every tested level across $A\in\{0.3,0.5,0.7\}$, so the degradation and recovery findings are not an artifact of the fixed $A=0.5$ choice. Both models also degrade as $A$ rises, brighter veiling being consistently more harmful, but the fog-aware model is markedly \emph{less} sensitive to $A$: at $i=2$ it is nearly flat between $A=0.3$ and $A=0.5$ before declining at $A=0.7$, whereas the baseline falls steadily across the range (Figure \ref{fig:sensitivity}). This reduced sensitivity is observed on a three-level grid from single representative runs, without the seed-level uncertainty of the primary results, so it is reported as a supporting trend rather than an established mechanism. What the analysis establishes firmly is that the fog-aware advantage holds in both direction and magnitude across the tested range of $A$.

\subsection{Limitations, Threats to Validity, and Alternative Explanations}
The study's main strength is internal validity: fog severity is the single manipulated variable, scene content and ground-truth annotations are identical across all conditions, and architecture, schedule, and hyperparameters are held fixed, so observed differences are attributable to visibility and to training-data composition rather than to confounds. The deterministic, resolution-dependent fog protocol, fixed seeds, public code, and single documented hardware configuration give the work strong \emph{reproducibility}.

The principal threat is \emph{construct validity}, the realism of the fog itself. The radial-distance prior is a heuristic degradation protocol, not a physically accurate depth map, and its weak empirical relationship with target geometry (Pearson $r=-0.085$, $r^2=0.007$) confirms it does not capture true scene depth. An alternative reading of the collapse is therefore that it partly reflects the prior's geometry rather than fog density per se. This is unlikely to drive the main result: the prior applies its \emph{strongest} attenuation at the image center, where the large targets cluster, yet those targets remain the most fog-robust through $i=7$ before collapsing only at the heaviest levels. Fog density, not spatial placement, is therefore the dominant driver, and the large-target robustness is if anything conservative, achieved despite the heaviest synthetic fog. The prior's central bias is nonetheless the most serious limitation for the size-stratified analysis specifically, and is carried into the conclusion as a qualification on any size-dependent claim.

A second, unexamined source of variation is scene illumination. The Anti-UAV300 RGB subset contains both day and night sequences (Section\ref{dataset}), but results are pooled across lighting conditions without stratification. This matters because the fog model attenuates scene radiance towards a fixed grey atmospheric light ($A=0.5$), an assumption better matched to daylight than to low-light night frames, where contrast and sensor-noise characteristics differ. Pooling across day and night may average over two distinct degradation behaviors, and the atmospheric light sensitivity analysis, which shows both models sensitive to A, suggests lighting could interact with fog severity. 
Day/night stratification was not conducted in the present work in order to preserve analytical depth on the primary fog-severity question. Stratifying by illumination would reduce the per-condition sample below the 91 test sequences used here, and an uneven day/night split, which could leave one lighting condition with too few sequences for reliable mAP trends across all eleven severity levels, two models and three seeds.

A further limitation is the training configuration. Hyperparameters follow Ultralytics defaults rather than a tuned schedule. Because both models share this configuration it does not confound their comparison but the absolute mAP values and the absolute size of the fog collapse, should be read as specific to this untuned setting. A tuned, target-adapted configuration could shift the degradation levels even if the threshold behavior persists. 

\emph{Generalizability} is bounded on several axes: a single dataset (Anti-UAV300 RGB), a single object class (UAV), a single architecture (YOLOv5m, chosen for direct comparability with Munir et al. \cite{munir_impact_2024}), the RGB modality only with the thermal stream excluded, and synthetic rather than real fog. The results should not be read as detector-, dataset-, or weather-general; they characterize one detector under one controlled, synthetic degradation.

\emph{Reliability} of the size-stratified and sensitivity findings is limited by the $27$-instance tiny sample, merged into the small category, and by the restricted three-level $\times$ three-$A$ sensitivity grid.

\emph{Scalability} of the pipeline is favorable: post-hoc fog generation and the deterministic prior extend cleanly to other clear datasets, which is what makes the protocol reusable as a benchmark.

These limitations define several directions for future work: validating
the identified visibility threshold on real fog imagery and depth-correct
synthetic fog, conducting day/night-stratified evaluation, and retraining
at native resolution to resolve the remaining resolution confound.
Generalizability should also be examined using thermal and RGB--T fusion,
alternative detector and tracker architectures, and independent anti-UAV
datasets. Multi-view tracking represents another potentially useful
extension; Elchik et al.~\cite{elchik2025multiview}, for example, developed
a stereo multi-view framework that integrates single-view trackers such as
FairMOT and YOLOv8 to address difficult small-object tracking scenarios.
Adapting similar multi-view or multi-sensor strategies to anti-UAV
detection could help determine whether complementary visual observations
increase robustness when visibility in an individual stream deteriorates.

\subsection{Ethical Considerations}
Anti-UAV detection is inherently dual-use. Improving detector reliability under poor visibility serves legitimate safety-critical aims, protecting airspace around airports and crowds of the kind disrupted at Gatwick \cite{hudson_gatwick_nodate}, but the same capability underpins anti-UAV systems whose deployment raises questions of privacy, proportionality, and automated targeting. This tension is specific to the present contribution: the same fog robustness that keeps protective systems operational in poor visibility also erodes a natural limit on persistent surveillance, since adverse weather currently degrades tracking. The use of synthetic fog is itself a modest ethical benefit, reducing the need to collect real surveillance footage under adverse conditions, though it also means claims of operational readiness must be tempered until validated on real data. This work is positioned as a controlled robustness benchmark rather than a deployable system, and any downstream use should be subject to the legal and oversight constraints governing anti-UAV operations.

\section{Conclusion}
\label{sec:conclusion}

Anti-UAV systems must operate in poor visibility, but current benchmarks use only clear-sky data, and previous work treats fog as simply present or absent. As a result, the effects of varying fog severity on the ground-to-air UAV detection remain poorly understood.

This study evaluated how increasing fog severity affects YOLOv5m detection, and whether synthetic fog augmentation provides robustness compared to a clear-sky baseline. Detection loss is front-loaded: most degradation occurs at low-to-moderate severity, with a $96\%$ reduction in mAP@0.5:0.95 from clear to heaviest fog, driven by lost recall and confidence rather than localization error. A uniform-fog control confirms this pattern reflects fog density rather than the fog protocol's spatial structure. Large targets remain the most robust, while the effect on small targets is inconclusive. Fog-aware training improves detection at all fog levels (up to $+0.320$ mAP@0.5:0.95 at $i=4$), raising the threshold for reliable detection but not preventing failure at the highest severity ($0.093$ at $i=9$). This improvement comes at a clear-sky cost of only $9.1\%$.

These findings are based on synthetic fog, a single RGB dataset, a single object class, and a single model architecture, so generalization is uncertain. The main limitations are the realism of the fog simulation and unresolved confounding between small-target scarcity and mismatched train/test resolutions. Size-dependent results should therefore be viewed as suggestive rather than definitive.

This work introduces the first severity-controlled fog benchmark for ground-to-air anti-UAV detection. The central practical implication is that reliability collapses within a narrow visibility band, so simple, clear vs adverse evaluations underestimate operational risk. Severity-controlled testing is needed to identify critical thresholds.

The next step is to validate these findings on real fog imagery, to see if the results transfer beyond synthetic conditions. Further improvements would involve using depth-corrected synthetic fog, training at native high resolutions, and extending the study to thermal data and alternative model architectures.

\bibliographystyle{ACM-Reference-Format}
\bibliography{references}

\appendix

\section{Methodology Visualization}
\label{sec:apx:methodology_vis}
\begin{figure}[H]
    \centering
    \includegraphics[width=1\linewidth]{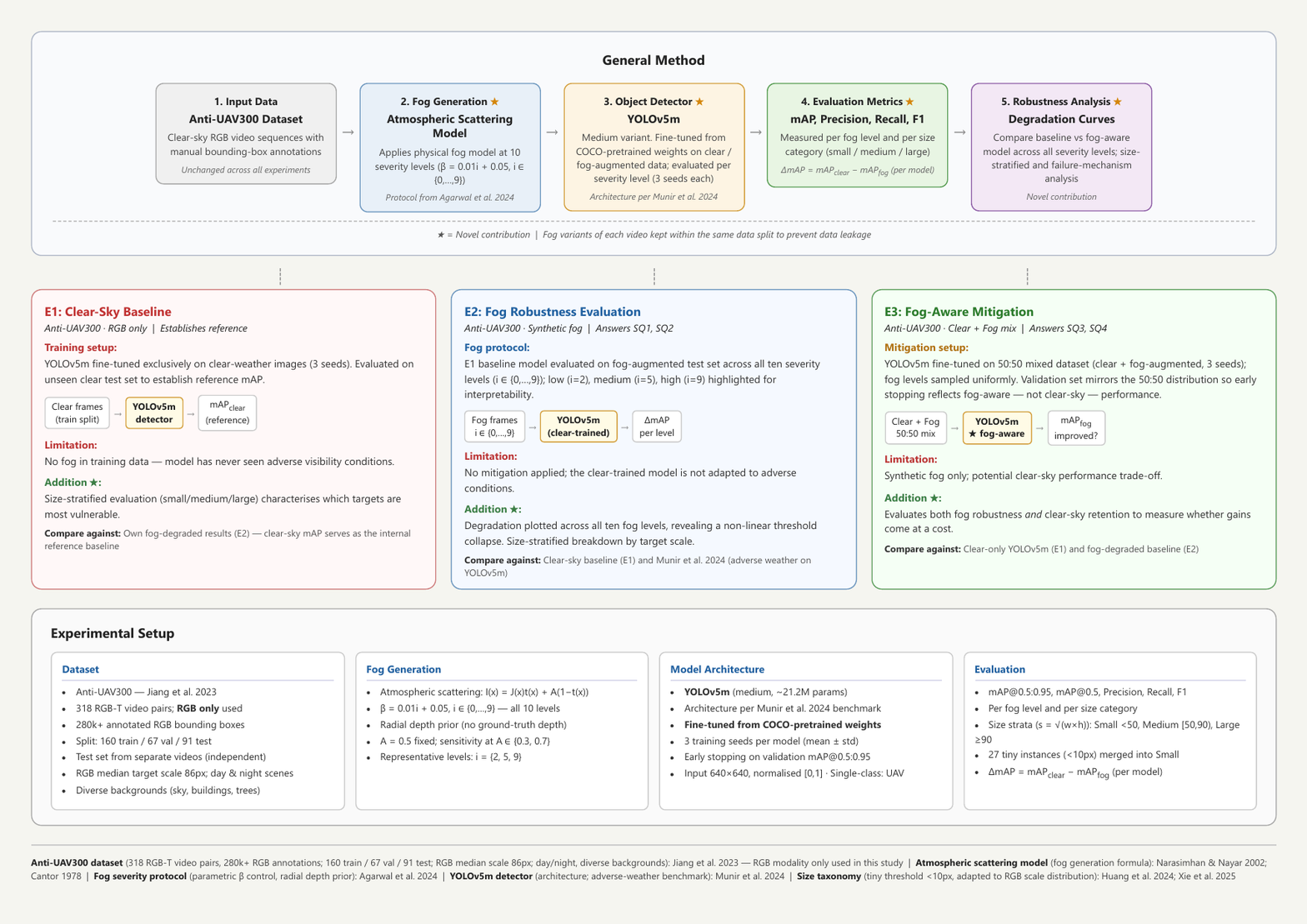}
    \caption{Methodology visualization showing the five-stage experimental pipeline, three experimental conditions (clear-sky baseline, fog robustness evaluation, and fog-aware mitigation), and key experimental setup parameters. \textit{Star} denotes novel contributions.}
    \label{fig:methods}
\end{figure}

\section{Data Configuration}
\label{sec:apx:data_conf}
\begin{table}[H]
    \centering
    \begin{tabular}{ll}
    \toprule
    \textbf{Split} & \textbf{Data} \\
    \midrule
    Training   & Clear images \\
    Validation & Clear images \\
    Test       & Clear images + fog (all severity levels) \\
    \bottomrule
    \end{tabular}
    \caption{Data configuration for the clear-sky baseline condition.}
\label{tab:baseline_splits}
\end{table}
\section{Fog Generation}
\label{sec:apx:fog_generation}
To evaluate the robustness of drone detection under degraded visibility conditions, synthetic fog is applied to the clear dataset using a physically-based atmospheric scattering model. The fog simulation follows the formulation introduced in prior work on visibility degradation\cite{agarwal_warlearn_2024}.

Foggy images are generated according to the standard atmospheric scattering model \cite{cantor_optics_1978, narasimhan_vision_2002}:
\begin{equation}
    I(x) = J(x) \cdot t(x) + A \cdot (1 - t(x))
\end{equation}
where $I(x)$ is the resulting foggy image, $J(x)$ the original clear image, $A$ the global atmospheric light, and $t(x)$ the medium transmission map. \noindent The transmission map is defined as:
\begin{equation}
    t(x) = e^{-\beta d(x)}
\end{equation}
where $\beta$ is the atmospheric scattering coefficient controlling fog density, and $d(x)$ the scene depth at pixel $x$.

This formulation models the reduction of scene radiance as light propagates through a participating medium such as fog. As \(\beta\) increases, transmission decreases and the image becomes increasingly washed out. This model is consistent with the approach of Agarwal et al.\cite{agarwal_warlearn_2024}

Since true per-pixel depth information is not available in the dataset, depth is approximated using a radial distance prior as proposed in Agarwal et al. \cite{agarwal_warlearn_2024}:
\begin{equation}
d(x)=-0.04\rho + \sqrt{\max(rows,cols)}    
\end{equation}

where $\rho$ is the Euclidean distance of pixel $x$ from the image center, and $rows$, $cols$ denote image height and width.

To create multiple controlled fog conditions, the scattering coefficient is varied according to:
\begin{equation}
\beta = 0.01i+0.05,\quad i \in \{0,1,\ldots,9\}
\end{equation}

In this implementation, each image is normalized to $[0,1]$ before the fog model is applied, so that the atmospheric light ($A=0.5$) and pixel values share a common scale; the result is rescaled to the original $8$-bit range afterwards. A single transmission map $t(x)$ is computed per image from the depth approximation above, clipped to $[0,1]$, and applied identically across all three color channels, with the atmospheric light added per channel.

This produces 10 discrete fog severity levels per image \cite{agarwal_warlearn_2024}. The atmospheric light parameter is fixed at \(A=0.5\), following the implementation of Agarwal et al. \cite{agarwal_warlearn_2024}.

\section{Baseline YOLOv5m Detection Performance}
\label{sec:apx:baseline_results}

\begin{table}[H]
\centering
\caption{Baseline YOLOv5m test-set performance across clear conditions and
all ten fog severity levels ($\beta = 0.01i + 0.05$, $A = 0.5$), reported
as mean $\pm$ standard deviation across three training seeds. The model is
trained on clear data only. From clear to the heaviest fog,
mAP@0.5:0.95 falls from $0.567$ to $0.022$ ($96.0\%$), while mAP@0.5
falls from $0.956$ to $0.048$ ($95.0\%$), showing that the collapse is
not an artefact of the stricter evaluation criterion.}
\label{tab:baseline_degradation}

\footnotesize
\setlength{\tabcolsep}{2pt}
\renewcommand{\arraystretch}{1.08}

\resizebox{\columnwidth}{!}{%
\begin{tabular}{llccccc}
\toprule
\textbf{Condition} &
$i$ &
\textbf{P} &
\textbf{R} &
\textbf{F1} &
\shortstack{\textbf{mAP}\\\textbf{@0.5}} &
\shortstack{\textbf{mAP}\\\textbf{@0.5:0.95}} \\
\midrule

Clear & -- &
$0.971 \pm 0.003$ &
$0.919 \pm 0.003$ &
$0.944 \pm 0.002$ &
$0.956 \pm 0.001$ &
$0.567 \pm 0.002$ \\

\midrule

Fog & 0 &
$0.866 \pm 0.023$ &
$0.745 \pm 0.028$ &
$0.801 \pm 0.018$ &
$0.819 \pm 0.020$ &
$0.373 \pm 0.017$ \\

Fog & 1 &
$0.800 \pm 0.031$ &
$0.580 \pm 0.038$ &
$0.672 \pm 0.028$ &
$0.679 \pm 0.032$ &
$0.278 \pm 0.020$ \\

Fog & 2 &
$0.727 \pm 0.025$ &
$0.382 \pm 0.028$ &
$0.501 \pm 0.028$ &
$0.503 \pm 0.032$ &
$0.188 \pm 0.017$ \\

Fog & 3 &
$0.339 \pm 0.025$ &
$0.362 \pm 0.030$ &
$0.349 \pm 0.011$ &
$0.336 \pm 0.017$ &
$0.122 \pm 0.008$ \\

Fog & 4 &
$0.320 \pm 0.042$ &
$0.187 \pm 0.008$ &
$0.235 \pm 0.007$ &
$0.229 \pm 0.023$ &
$0.088 \pm 0.009$ \\

Fog & 5 &
$0.258 \pm 0.043$ &
$0.083 \pm 0.001$ &
$0.125 \pm 0.004$ &
$0.155 \pm 0.024$ &
$0.066 \pm 0.010$ \\

Fog & 6 &
$0.181 \pm 0.020$ &
$0.042 \pm 0.002$ &
$0.068 \pm 0.002$ &
$0.104 \pm 0.010$ &
$0.047 \pm 0.006$ \\

Fog & 7 &
$0.146 \pm 0.025$ &
$0.026 \pm 0.001$ &
$0.043 \pm 0.002$ &
$0.082 \pm 0.013$ &
$0.038 \pm 0.006$ \\

Fog & 8 &
$0.113 \pm 0.026$ &
$0.016 \pm 0.001$ &
$0.028 \pm 0.002$ &
$0.063 \pm 0.013$ &
$0.029 \pm 0.005$ \\

Fog & 9 &
$0.085 \pm 0.012$ &
$0.010 \pm 0.000$ &
$0.018 \pm 0.000$ &
$0.048 \pm 0.006$ &
$0.022 \pm 0.003$ \\

\bottomrule
\end{tabular}%
}

\end{table}

\section{Fog-aware YOLOv5m Detection Performance}
\label{sec:apx:fogaware_results}

\begin{table}[H]
\centering
\caption{Fog-aware YOLOv5m test-set performance across clear conditions and
all ten fog severity levels ($A = 0.5$), reported as mean $\pm$ standard
deviation across three training seeds. The model is trained and validated on
a 50:50 clear/fog mixture, with $i \in \{0,\ldots,9\}$. At every fog level,
the model outperforms the clear-trained baseline; the largest gain is
$+0.320$ mAP@0.5:0.95 at $i=4$.}
\label{tab:fogaware_degradation}

\footnotesize
\setlength{\tabcolsep}{2pt}
\renewcommand{\arraystretch}{1.08}

\resizebox{\columnwidth}{!}{%
\begin{tabular}{llccccc}
\toprule
\textbf{Condition} &
$i$ &
\textbf{P} &
\textbf{R} &
\textbf{F1} &
\shortstack{\textbf{mAP}\\\textbf{@0.5}} &
\shortstack{\textbf{mAP}\\\textbf{@0.5:0.95}} \\
\midrule

Clear & -- &
$0.946 \pm 0.012$ &
$0.896 \pm 0.009$ &
$0.920 \pm 0.009$ &
$0.938 \pm 0.009$ &
$0.515 \pm 0.006$ \\

\midrule

Fog & 0 &
$0.942 \pm 0.008$ &
$0.862 \pm 0.006$ &
$0.901 \pm 0.003$ &
$0.915 \pm 0.004$ &
$0.496 \pm 0.001$ \\

Fog & 1 &
$0.935 \pm 0.001$ &
$0.844 \pm 0.004$ &
$0.888 \pm 0.002$ &
$0.900 \pm 0.003$ &
$0.482 \pm 0.002$ \\

Fog & 2 &
$0.926 \pm 0.008$ &
$0.816 \pm 0.005$ &
$0.867 \pm 0.001$ &
$0.882 \pm 0.003$ &
$0.463 \pm 0.004$ \\

Fog & 3 &
$0.919 \pm 0.005$ &
$0.790 \pm 0.003$ &
$0.850 \pm 0.002$ &
$0.860 \pm 0.003$ &
$0.437 \pm 0.005$ \\

Fog & 4 &
$0.900 \pm 0.010$ &
$0.760 \pm 0.002$ &
$0.824 \pm 0.003$ &
$0.829 \pm 0.004$ &
$0.408 \pm 0.006$ \\

Fog & 5 &
$0.869 \pm 0.014$ &
$0.709 \pm 0.004$ &
$0.781 \pm 0.006$ &
$0.780 \pm 0.005$ &
$0.365 \pm 0.003$ \\

Fog & 6 &
$0.805 \pm 0.009$ &
$0.653 \pm 0.010$ &
$0.721 \pm 0.006$ &
$0.703 \pm 0.007$ &
$0.314 \pm 0.004$ \\

Fog & 7 &
$0.741 \pm 0.007$ &
$0.536 \pm 0.003$ &
$0.622 \pm 0.004$ &
$0.566 \pm 0.007$ &
$0.243 \pm 0.002$ \\

Fog & 8 &
$0.645 \pm 0.013$ &
$0.399 \pm 0.004$ &
$0.493 \pm 0.006$ &
$0.401 \pm 0.010$ &
$0.165 \pm 0.003$ \\

Fog & 9 &
$0.517 \pm 0.014$ &
$0.251 \pm 0.005$ &
$0.338 \pm 0.007$ &
$0.237 \pm 0.007$ &
$0.093 \pm 0.002$ \\

\bottomrule
\end{tabular}%
}

\end{table}

\section{Degradation Curve}
\label{sec:apx:deg_curve}
\begin{figure}[H]
    \centering
    \includegraphics[width=1\linewidth]{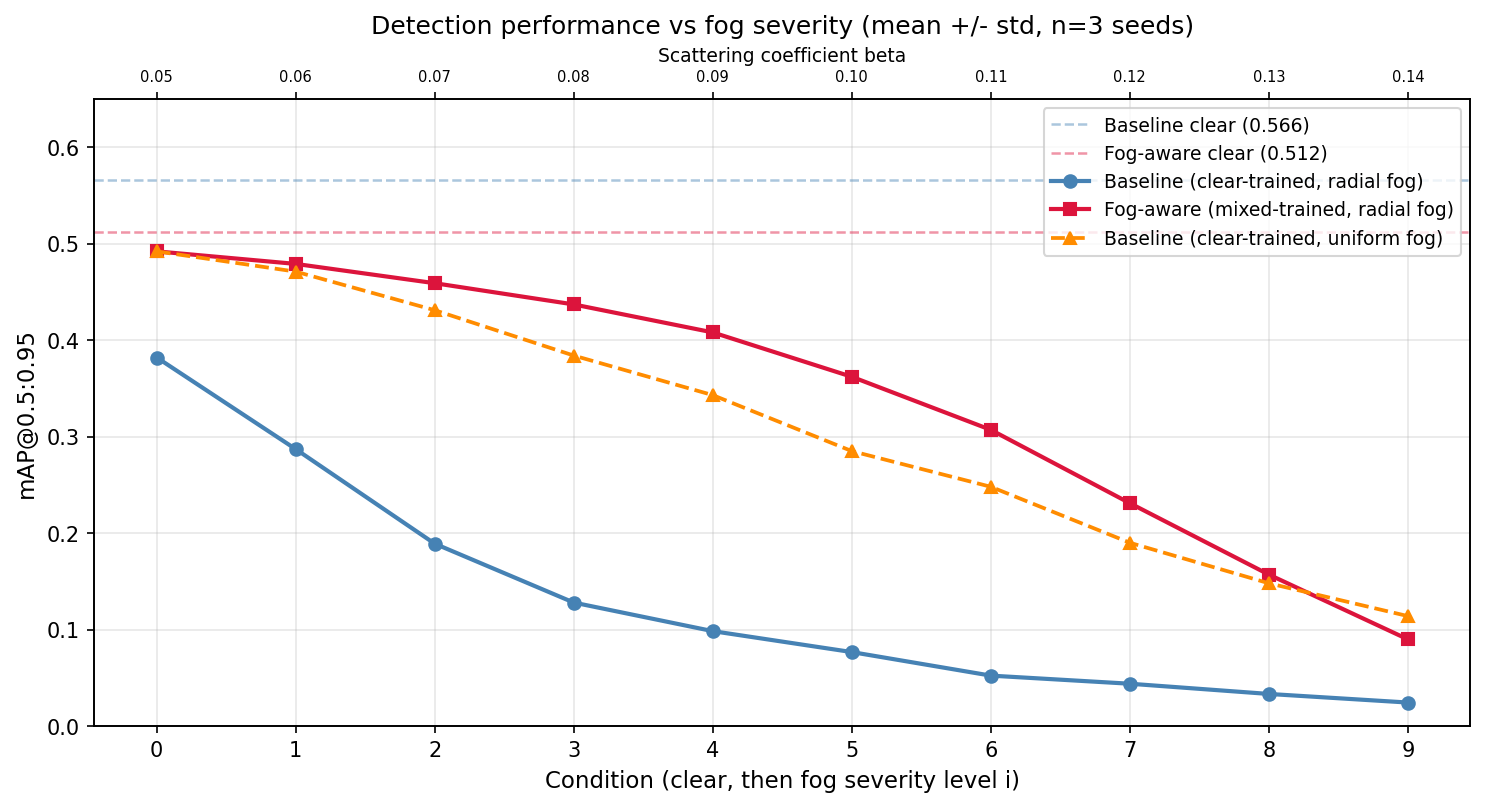}
    \caption{Detection performance of the baseline and fog-aware YOLOv5m models on the test set across all ten fog severity levels ($A=0.5$). Yellow line represents the uniform-fog control tested at only \textbf{ 1 seed run}, trained using baseline weights.}
\label{fig:degradation}
\end{figure}

\section{Absolute Performance Loss}
\label{sec:apx:delta_map}
\begin{figure}[H]
    \centering
    \includegraphics[width=1\linewidth]{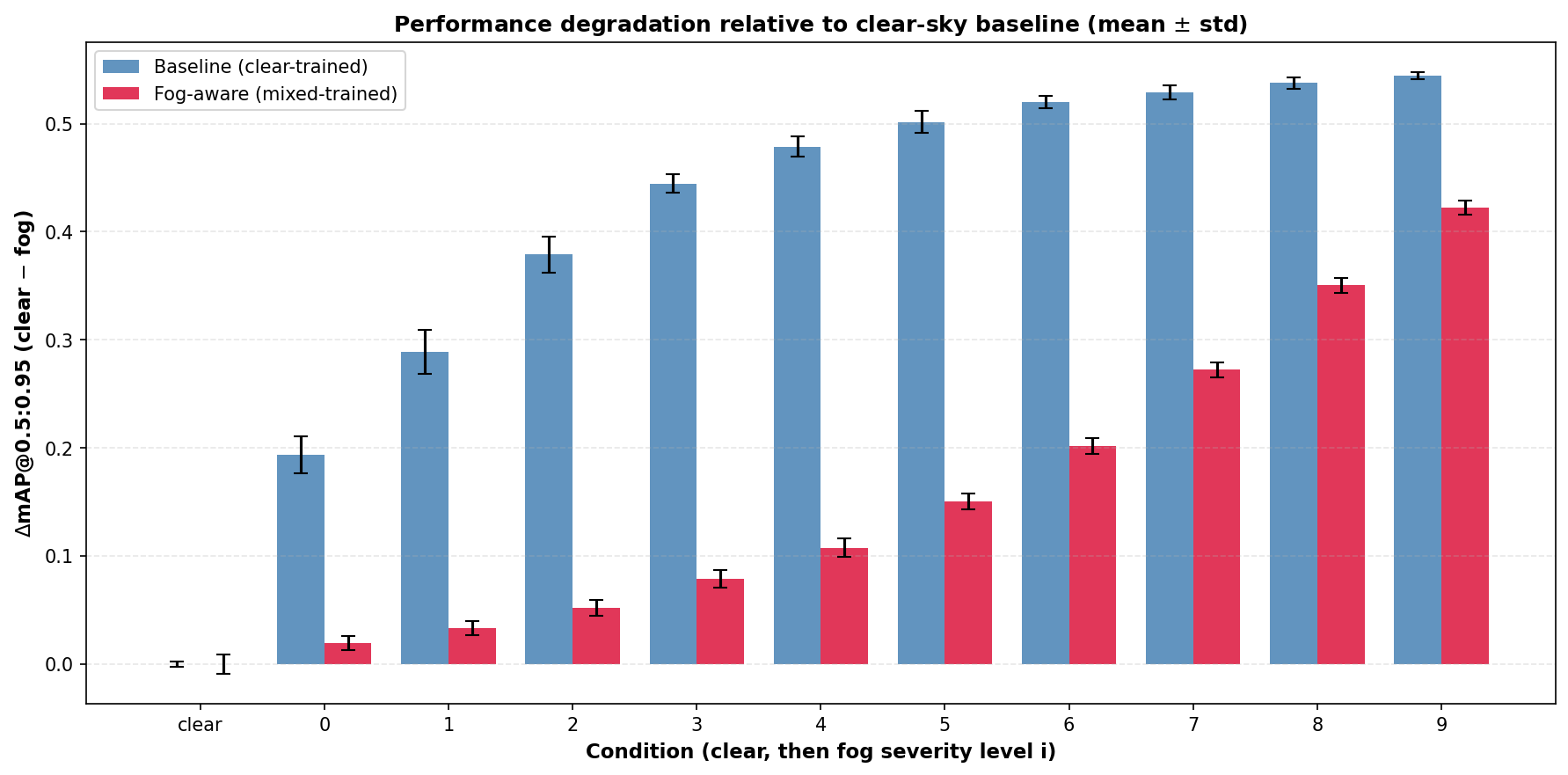}
    \caption{Absolute performance loss ($\Delta$mAP@0.5:0.95) relative to each model's clear-sky baseline across all ten fog severity levels ($A=0.5$).}
\label{fig:delta_map}
\end{figure}

\section{Size-Stratified Analysis}
\label{sec:apx:size-strat}
\begin{table}[H]
\centering
\begin{tabular}{lccc|ccc}
\toprule
& \multicolumn{3}{c|}{\textbf{Baseline}} & \multicolumn{3}{c}{\textbf{Fog-aware}} \\
\textbf{Severity} & Small & Medium & Large & Small & Medium & Large \\
\midrule
Clear   & 0.528 & 0.525 & 0.596 & 0.448 & 0.488 & 0.550 \\
$i{=}0$ & 0.251 & 0.334 & 0.452 & 0.415 & 0.455 & 0.548 \\
$i{=}1$ & 0.195 & 0.248 & 0.353 & 0.397 & 0.441 & 0.540 \\
$i{=}2$ & 0.153 & 0.165 & 0.237 & 0.371 & 0.418 & 0.528 \\
$i{=}3$ & 0.107 & 0.093 & 0.156 & 0.351 & 0.386 & 0.509 \\
$i{=}4$ & 0.070 & 0.049 & 0.113 & 0.300 & 0.351 & 0.495 \\
$i{=}5$ & 0.051 & 0.027 & 0.086 & 0.242 & 0.309 & 0.453 \\
$i{=}6$ & 0.040 & 0.015 & 0.063 & 0.174 & 0.257 & 0.406 \\
$i{=}7$ & 0.035 & 0.012 & 0.040 & 0.117 & 0.200 & 0.323 \\
$i{=}8$ & 0.026 & 0.009 & 0.024 & 0.086 & 0.137 & 0.217 \\
$i{=}9$ & 0.021 & 0.009 & 0.015 & 0.059 & 0.075 & 0.127 \\
\bottomrule
\end{tabular}
\caption{Size-stratified detection performance (mAP@0.5:0.95) on the test set across clear conditions and all ten fog severity levels for the baseline and fog-aware models. Targets are grouped Small/Medium/Large; the 27 tiny instances are merged into Small (Section \ref{dataset}). $A=0.5$.}
\label{tab:size_results}
\end{table}

\section{Resolution Analysis}
\label{sec:apx:resolution}
\begin{table}[H]
\centering
\begin{tabular}{lcccc}
\toprule
& \multicolumn{2}{c}{\textbf{Recall}} & \multicolumn{2}{c}{\textbf{mAP@0.5:0.95}} \\
\textbf{Severity} & 640 & 1280 & 640 & 1280 \\
\midrule
Clear   & 0.947 & 0.917 & 0.528 & 0.517 \\
\midrule
$i{=}0$ & 0.824 & 0.743 & 0.251 & 0.208 \\
$i{=}1$ & 0.715 & 0.546 & 0.195 & 0.140 \\
$i{=}2$ & 0.624 & 0.517 & 0.153 & 0.093 \\
$i{=}3$ & 0.374 & 0.304 & 0.107 & 0.060 \\
$i{=}4$ & 0.213 & 0.136 & 0.070 & 0.040 \\
$i{=}5$ & 0.113 & 0.079 & 0.051 & 0.029 \\
$i{=}6$ & 0.083 & 0.049 & 0.040 & 0.021 \\
$i{=}7$ & 0.056 & 0.029 & 0.035 & 0.016 \\
$i{=}8$ & 0.043 & 0.020 & 0.026 & 0.014 \\
$i{=}9$ & 0.028 & 0.016 & 0.021 & 0.013 \\
\bottomrule
\end{tabular}
\caption{Baseline small-target recall and mAP@0.5:0.95 at $640$ vs $1280$ inference resolution across clear conditions and all ten fog severity levels (model trained at $640 \times 640$, $A=0.5$). Inference at $1280$ is lower than at $640$ at every condition.}
\label{tab:resolution}
\end{table}
\section{Confidence Distribution}
\label{sec:apx:conf_dist}
\begin{figure}[H]
    \centering
    \includegraphics[width=\linewidth]{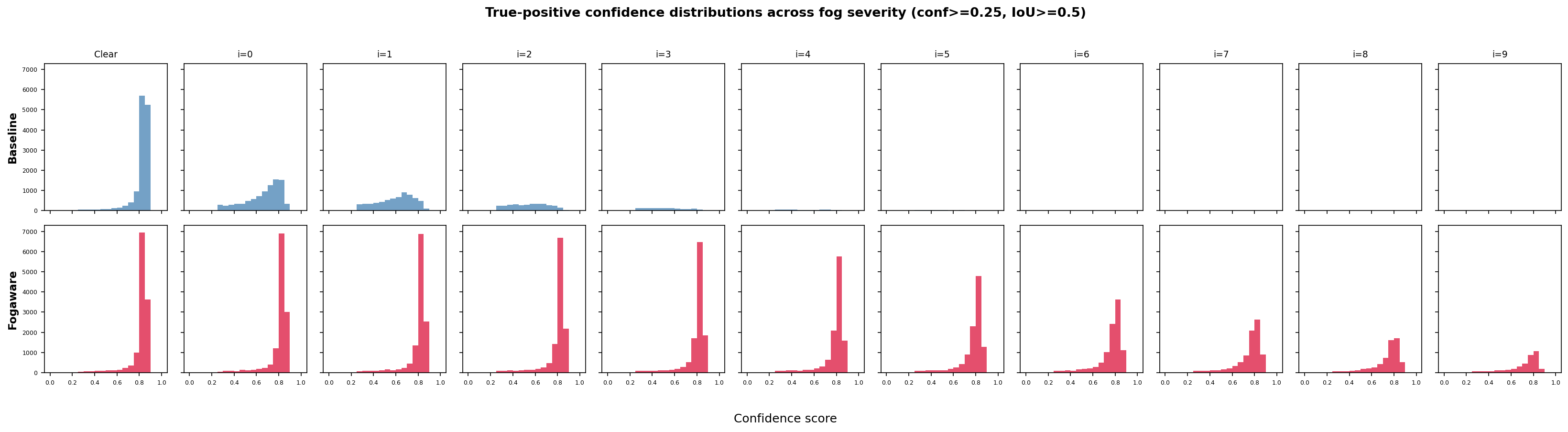}
    \caption{True-positive confidence distributions across fog severity (conf$\geq0.25$, IoU$\geq0.5$; single representative run per model). The baseline's confident true positives fall to near-zero from $i\approx4$ (empty panels reflect detection loss, not missing data); the fog-aware model retains a peak near $0.8$ at every level.}
\label{fig:conf_dist}
\end{figure}

\section{False Discovery Rate}
\label{sec:apx:fp_rate}
\begin{figure}[H]
    \centering
    \includegraphics[width=\linewidth]{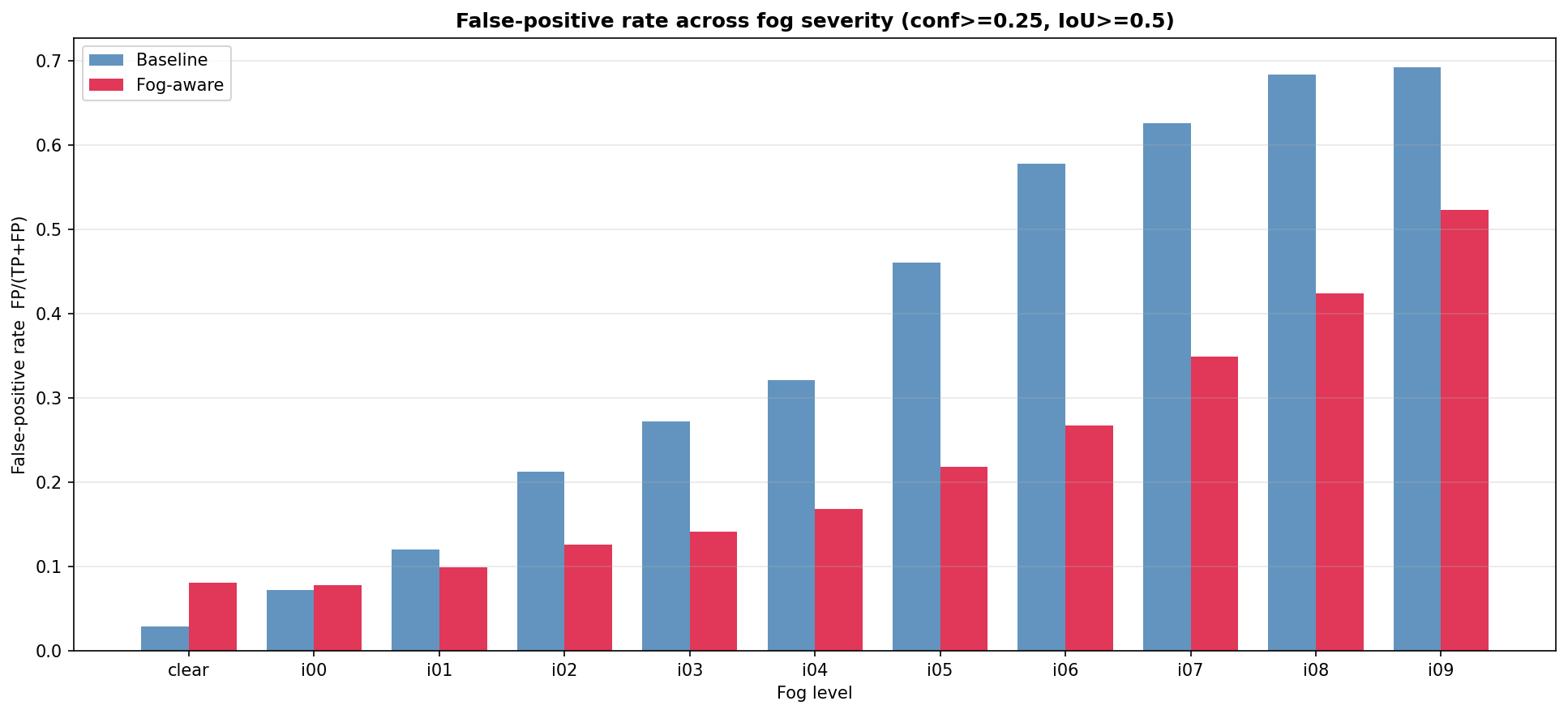}
    \caption{False-discovery rate (FP/(TP+FP), i.e.\ $1-$precision) across fog severity (conf$\geq0.25$, IoU$\geq0.5$; single representative run). Both models rise with fog; the fog-aware model stays lower from $i=2$ on, at the cost of a marginally higher rate under clear conditions.}
\label{fig:fp_rate}
\end{figure}

\section{IoU Stability}
\label{sec:apx:iou_stab}
\begin{figure}[H]
    \centering
    \includegraphics[width=\linewidth]{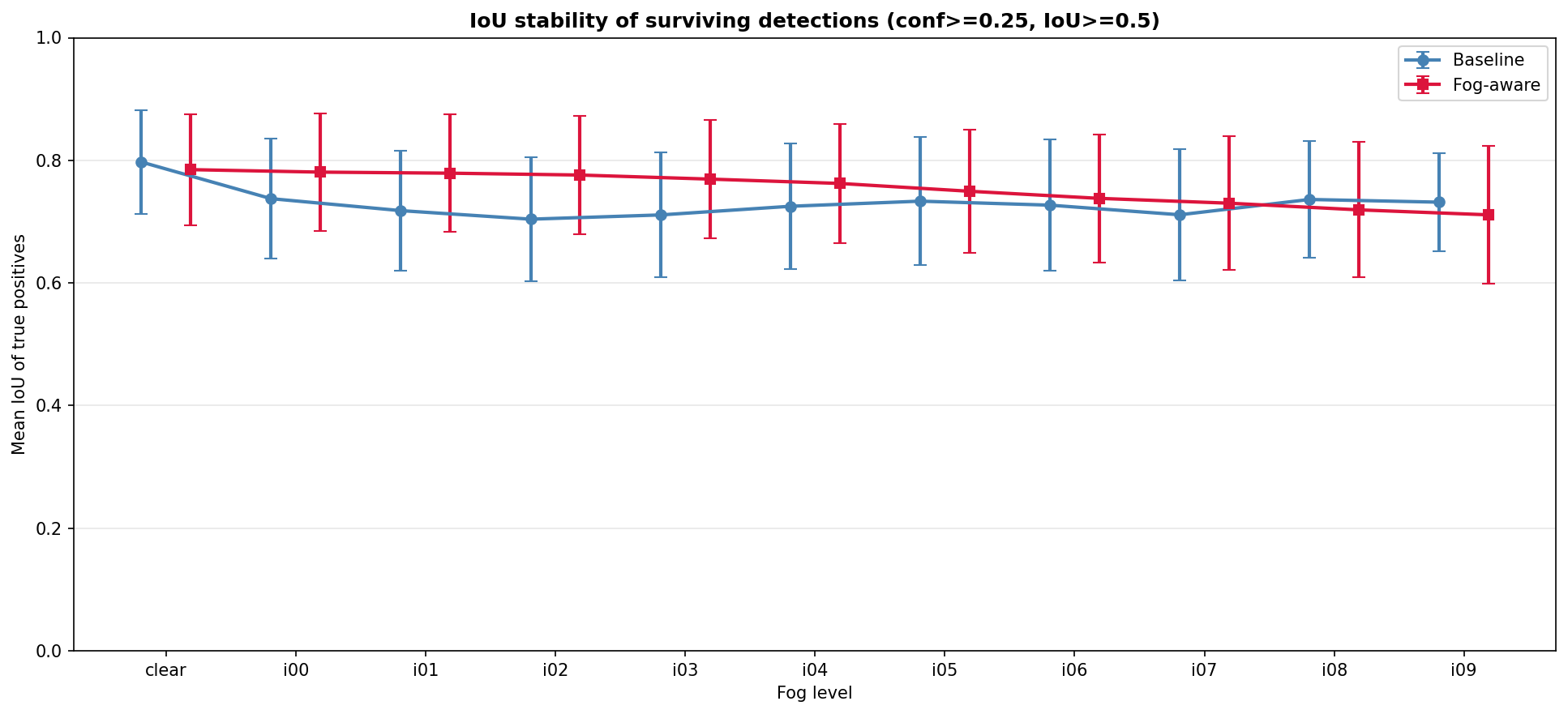}
    \caption{Mean IoU of surviving true positives across fog severity (conf $\geq0.25$, IoU $\geq0.5$; single representative run; error bars $\pm1$ std). Localization quality holds at $\sim0.75$ for both models at every level: surviving detections remain well-localized even as recall collapses.}
\label{fig:iou_stab}
\end{figure}

\section{Precision-Recall Curves}
\label{sec:apx:pr_curves}
\begin{figure}[H]
    \centering
    \includegraphics[width=\linewidth]{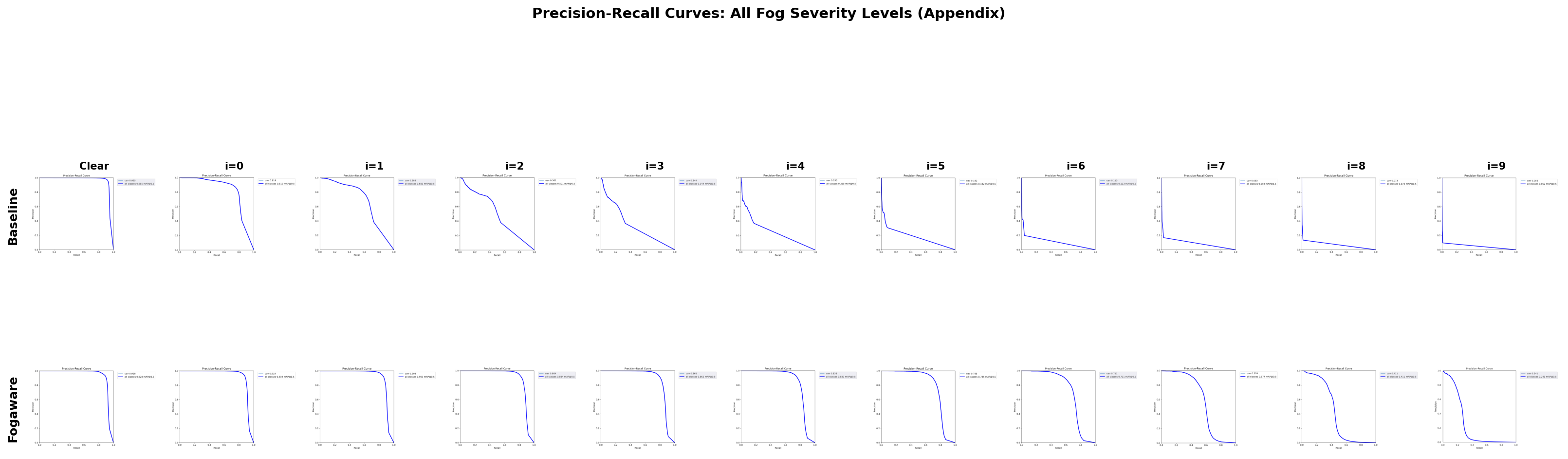}
    \caption{Precision--recall curves for the baseline and fog-aware models across clear and all ten fog severity levels (single representative run per model). Baseline curves contract toward the origin as fog density rises; the fog-aware model retains substantially more area at every level.}
\label{fig:pr_curves}
\end{figure}

\FloatBarrier

\clearpage

\end{document}